\documentclass[10pt,journal,compsoc]{IEEEtran}
\usepackage{amsmath,amsfonts}
\usepackage{algorithmic}
\usepackage{array}
\usepackage[caption=false,font=normalsize,labelfont=sf,textfont=sf]{subfig}
\usepackage{textcomp}
\usepackage{stfloats}
\usepackage{url}
\usepackage{verbatim}
\usepackage{graphicx}
\usepackage{cite}

\usepackage{multicol}
\usepackage{multirow}
\usepackage{epstopdf}
\usepackage[linesnumbered,ruled,vlined]{algorithm2e}
\usepackage{makecell}
\usepackage{amsthm}

\theoremstyle{definition}
\newtheorem{definition}{Definition}
\newtheorem{assumption}{Assumption}

\newtheorem{theorem}{Theorem}

\newtheorem{corollary}{Corollary}

\begin{document}

\title{Robust Model Aggregation for Heterogeneous Federated Learning: Analysis and Optimizations}

\author{Yumeng Shao, Jun Li, Long Shi, Kang Wei, Ming Ding, Qianmu Li, Zengxiang Li, Wen Chen, and Shi Jin
\thanks{Y. Shao, J. Li, L. Shi, and K. Wei are with the School of Electronic and Optical Engineering, Nanjing University of Science and Technology, Nanjing 210094, China. K.Wei is now with the Department of Computing, The Hong Kong Polytechnic University, Hong Kong, China (e-mail: \{shaoyumeng, jun.li, kang.wei\}@njust.edu.cn, slong1007@gmail.com).}
\thanks{ M. Ding is with Data61, CSIRO, Sydney, Australia (e-mail: ming.ding@data61.csiro.au).}
\thanks{ Q. Li is with the Digital Economy Research Institute, Nanjing University of Science and Technology, Nanjing 210094, Jiangsu, China. (e-mail: qianmu@njust.edu.cn).}
\thanks{Z. Li is with the Institute of High Performance Computing, A*STAR, Singapore. He is also with the Digital Research Institute, ENN Group, China (e-mail: lizengxiang@enn.cn).}
\thanks{W. Chen is with the Department of Electronic Engineering, Shanghai Jiao Tong University, Shanghai 200240, China (e-mail: wenchen@sjtu.edu.cn). }
\thanks{S. Jin is with the National Mobile Communications Research Laboratory, Southeast University, Nanjing 210096, China (e-mail: jinshi@seu.edu.cn). }
}

\markboth{}%
{Shell \MakeLowercase{\textit{et al.}}: Robust Aggregation for Heterogeneous Federated Learning: Performance Analysis and Optimizations}


\IEEEtitleabstractindextext{

\begin{abstract}
Conventional synchronous federated learning (SFL) frameworks suffer from performance degradation in heterogeneous systems due to imbalanced local data size and diverse computing power on the client side.
To address this problem, asynchronous FL (AFL) and semi-asynchronous FL have been proposed to recover the performance loss by allowing asynchronous aggregation.
However, asynchronous aggregation incurs a new problem of inconsistency between local updates and global updates.
Motivated by the issues of conventional SFL and AFL, we first propose a time-driven SFL (T-SFL) framework for heterogeneous systems.
The core idea of T-SFL is that the server aggregates the models from different clients, each with varying numbers of iterations, at regular time intervals.
To evaluate the learning performance of T-SFL, we provide an upper bound on the global loss function.
Further, we optimize the aggregation weights to minimize the developed upper bound.
Then, we develop a discriminative model selection (DMS) algorithm that removes local models from clients whose number of iterations falls below a predetermined threshold. In particular, this algorithm ensures that each client's aggregation weight accurately reflects its true contribution to the global model update, thereby improving the efficiency and robustness of the system.
To validate the effectiveness of T-SFL with the DMS algorithm, we conduct extensive experiments using several popular datasets including MNIST, Cifar-10, Fashion-MNIST, and SVHN. The experimental results demonstrate that T-SFL with the DMS algorithm can reduce the latency of conventional SFL by 50\%, while achieving an average 3\% improvement in learning accuracy over state-of-the-art AFL algorithms.
\end{abstract}

}
\maketitle

\IEEEdisplaynontitleabstractindextext
\IEEEpeerreviewmaketitle

\IEEEraisesectionheading{\section{Introduction}\label{sec:intro}}
\IEEEPARstart{W}{ith} the rapid advancement of the Internet of Things (IoT), the computing power of intelligent devices has been exploding at an unprecedented rate, allowing edge devices to execute a wide range of machine learning (ML) tasks \cite{9609994}. 
However, conventional ML frameworks with a centralized server cannot efficiently utilize the computing power of edge devices ~\cite{DBLP:conf/hicss/ZhangDFP21}.
To tackle this challenge, distributed machine learning (DML) architectures have been proposed to enable data processing at the edge in a distributed manner~\cite{9945975}.
In traditional DML, data exchange between various edge devices imposes a high burden of communication cost on every client \cite{8805879}.
As a novel DML framework, federated learning (FL) has manifested its promising advantages in reducing communication costs \cite{9999679}.
In FL, models are trained across distributed clients, each with a certain number of local training iterations, and then aggregated by a server \cite{DBLP:journals/wc/LiuPKINE20}.
Hence, FL can jointly accomplish machine learning tasks without transmitting raw data \cite{DBLP:journals/tifs/WeiLDMYFJQP20}.
Recently, FL has found widespread application in commercial practices, such as Tencent's WeBank~\cite{2020The}, Google's Gboard project~\cite{DBLP:journals/corr/abs-1912-01218}, and pharmaceutical labs like MELLODDY \cite{MELLODDY}.

Conventional FL involves updating the global model after all clients complete the same number of local training iterations, resulting in synchronous FL (SFL) \cite{DBLP:journals/tmc/WeiLDMSZP22, DBLP:journals/network/MaLDYSQP20, DBLP:journals/tccn/ChuLWWDZQC22, DBLP:journals/cim/MaLSDWHP22}.
Conventional SFL utilizes the \emph{FedAvg} algorithm to generate a high-quality global model by setting aggregation weights proportional to each client's data size \cite{DBLP:conf/aistats/McMahanMRHA17}.
Nevertheless, most real-world systems are heterogeneous, with varying data sizes and computing power across different devices \cite{9813696}.
In heterogeneous systems, SFL requires the server to wait until the last client completes the required number of local training iterations \cite{DBLP:journals/tist/YangLCT19, 10234675}.
To improve SFL's performance in heterogeneous systems, various aggregation laws have been proposed.
For example, \cite{DBLP:journals/corr/abs-2010-05958} suggests a heuristic weight aggregating algorithm, called \emph{FedAT}, to balance data size and aggregation weights.
Ref. \cite{DBLP:conf/icc/NishioY19} develops the \emph{FedCS} algorithm that aggregates local models based on each client's computing power.
The authors in \cite{9292461} propose an efficient synchronization algorithm \emph{ESync} to coordinate local parties with different iterations, and analyze the trade-off between convergence accuracy and communication efficiency.
Later on, \cite{DBLP:conf/aistats/RuanZLJ21} proposes an aggregation scheme that converges in heterogeneous systems, where devices may be inactive or upload incomplete updates.
Ref. \cite{DBLP:journals/tc/WuHLMMJ21} proposes an aggregation law to mitigate the influence of stragglers, crashes, and model staleness.
The authors in \cite{10316585} propose a resilient estimation approach to the locally trained models during aggregation, which applies a modified geometric mean aggregation over the local models' parameters.

The FL frameworks in the aforementioned works remain synchronous in essence and are susceptible to latency issues caused by synchronous aggregation in heterogeneous systems.
To address this limitation, recent works have proposed asynchronous FL (AFL) and semi-asynchronous FL frameworks \cite{DBLP:journals/corr/abs-2109-04269}.
In AFL, the global model update occurs as soon as the server receives any local update.
This allows clients to start their next local training without waiting for others.
However, the frequent global model updates can result in inconsistent model updates, as clients with slower training may fall behind the current global model, thereby impairing the learning performance of AFL \cite{DBLP:conf/bigdataconf/ChenNSR20}.
Semi-AFL is a compromise between SFL and AFL, where the global model update occurs after the server receives a predefined number of local updates \cite{DBLP:journals/corr/abs-2106-06639}.
By adjusting this number, semi-AFL can trade off synchronization and consistency for various scenarios \cite{DBLP:journals/corr/abs-2110-02177}.
In spite of this, it unavoidably inherits the weaknesses of both SFL and AFL.

Driven by these issues, we propose a time-driven synchronized federated learning (T-SFL) framework to reduce the latency of conventional SFL and address the inconsistency of AFL in heterogeneous systems.
In this framework, the entire training process is divided into multiple communication intervals in the time domain.
At the end of each interval, each client uploads the trained local model, which is then aggregated by the server.
This approach guarantees time-driven synchronization among all clients and reduces the latency compared to conventional SFL methods.
Particularly, since the computing power and data size may vary among clients, the number of local training iterations differs for each individual client at each specific interval.
The main contributions are listed as follows:

\begin{itemize}
        \item {The proposed T-SFL framework enables the global model aggregation from different clients with diverse numbers of local training iterations at periodic intervals. In contrast to SFL and AFL, the proposed framework decreases the latency and resolves the inconsistency in heterogeneous FL systems. }
        \item {We derive an upper bound on the loss function to assess the learning performance of T-SFL. Subsequently, we optimize aggregation weights to enhance the learning performance by minimizing the bound.  }
        \item {We propose a model aggregation algorithm, named discriminative model selection (DMS) algorithm, to ensure that the global aggregation weight correctly reflects the corresponding contribution of each client. }
        \item{The experimental results on the MNIST, Cifar-10, Fashion-MNIST, and SVHN datasets demonstrate that T-SFL with DMS not only reduces the training time of conventional SFL by 50\%, but also achieves better learning performance than other conventional aggregation rules such as \emph{FedAvg}, \emph{FedProx}, and \emph{FedAsync}. In particular, on the SVHN dataset, T-SFL with DMS improves learning accuracy by approximately 7\% compared to \emph{FedAsync}.}
\end{itemize}

The remainder of this paper is organized as follows.
Section \ref{sec_pre} introduces the background and related works of this paper.
Then we describe the T-SFL framework and analyze its performance in Section \ref{sec_sysmod}.
Afterwards, we propose the DMS algorithm in Section~\ref{sec_alg}.
The experimental results are presented in Section~\ref{sec_exp}.
Section~\ref{sec_con} concludes this paper.
In addition, Table~\ref{tab:summ_nota} lists the main notation used in this paper.

\begin{table}[ht]
\caption{Summary of main notation}
\centering
\begin{tabular}{l||l}
\hline
Notation& Describtion\\
\hline\hline
\hline
$T$ & Total communication round\\
\hline
$t$ & Index of communication round\\
\hline
$H$ & Maximum number of local iteration\\
\hline
$h$ & Index of local iteration\\
\hline
\multirow{2}*{$\boldsymbol{w}_{i}^{t}$} & The model parameters of the $i$-th client at \\
&the $t$-th communication interval\\
\hline
\multirow{2}*{$\boldsymbol{w}_{i,h}^{t}$} & The model parameters of the $i$-th client at the $t$-th \\
&communication interval, after local iteration $h$\\
\hline
$\boldsymbol{w}^{t}$ & The global model parameters at the $t$-th interval\\
\hline
\multirow{2}*{$\boldsymbol{g}^{t}_{i,h}$} & The stochastic gradient of the $i$-th client at \\
& the $t$-th communication interval, after local iteration $h$\\
\hline
\multirow{2}*{$\bar{\boldsymbol{g}}^{t}_{i,h}$} & The full batch gradient of the $i$-th client at \\
& the $t$-th communication interval, after local iteration $h$\\
\hline
\multirow{2}*{${\xi_{i,h}^t}$} & A mini-batch sampled from the $i$-th client at \\
& the $t$-th communication interval, after local iteration $h$\\
\hline
\multirow{2}*{$\rho^t_i$} & The aggregation weight of the $i$-th client \\
& at the $t$-th communication interval\\
\hline
$\Gamma_i$ & The non-IID degree of the $i$-th client\\
\hline
$\eta$ & The learning rate\\
\hline
$N$ & The number of clients\\
\hline
$\vert\mathcal{D}_i \vert$ & The data size of the $i$-th client \\
\hline
$F(\cdot)$ & The global loss function \\
\hline
$F_i(\cdot)$ & The local loss function of the $i$-th client \\
\hline
\end{tabular}
\label{tab:summ_nota}
\end{table}

\section{Preliminary}\label{sec_pre}
\subsection{Conventional Synchronous Federated Learning}
In an FL system, there are $N$ clients, with the $i$-th client possessing the dataset $\mathcal{D}_i$ of size $\vert\mathcal{D}_i \vert$, $i=1,\dots,N$. In conventional SFL, each client trains its local model (e.g., CNN network) for a predefined number of iterations and sends the trained model to the server.
Once the server collects the local models from all clients, it generates the global model.
The clients then update their local models with the global model.
This procedure is known as a communication round, and multiple communication rounds are performed in FL.

During the $t$-th communication round, the server performs global aggregation according to the aggregation rule, e.g., \emph{FedAvg}, $\boldsymbol{w}^{t}=\sum_{i=1}^N \frac{\vert \mathcal{D}_i \vert}{\sum_{i=1}^N \vert \mathcal{D}_i \vert} \boldsymbol{w}_{i}^{t}$, where $\boldsymbol{w}^{t}$ denotes the global model and $\boldsymbol{w}_{i}^{t}$ denotes the local model of the $i$-th client, respectively. The global loss function of FL is defined as $F(\boldsymbol{w}^t) = \frac{1}{N}\sum_{i=1}^N F_i(\boldsymbol{w}^t)$, where $F_i(\cdot)$ is local loss function of the $i$-th client. 
It is worth noting that a lower value of the loss function corresponds to a higher learning accuracy. 
Eventually, the FL system outputs the global model $\boldsymbol{w}^T$ that minimizes the global loss $F(\boldsymbol{w}^T)$, where $T$ is the total number of communication rounds.

\subsection{Asynchronous FL and Semi-asynchronous FL}
In AFL, the server updates the global model immediately after receiving any local model.
To improve the convergence of AFL, \cite{DBLP:journals/corr/abs-1903-03934} introduces the \emph{FedAsync} algorithm.
Ref. \cite{DBLP:conf/mlsys/LiSZSTS20} proposes the \emph{FedProx} algorithm to enhance the learning accuracy by incorporating the proximal item in the loss function.
Ref. \cite{DBLP:journals/tnn/ChenSJ20} suggests a temporally weighted aggregation rule to assign a higher weight to the most recent local models.
Ref. \cite{DBLP:conf/hpcc/ShiLWCY020} proposes an aggregation rule where the weight of each client decreases with increasing staleness value.
Ref. \cite{9831060} introduces a time-triggered FL to diminish latency for heterogeneous systems, but this approach still suffers from model inconsistency issues due to asynchronous updates.

Subsequently, several works have investigated semi-asynchronous FL frameworks as an intermediate between conventional SFL and AFL.
In semi-asynchronous FL, the server caches local models that arrive first and then aggregates them after collecting a certain number from the cached models.
In \cite{DBLP:journals/corr/abs-2106-06639}, the authors propose a novel buffered semi-asynchronous aggregation method, called \emph{FedBuff}, that establishes a private model buffer in the server. Moreover, \cite{DBLP:journals/corr/abs-2110-02177} develops a buffered semi-asynchronous secure aggregation method that has a higher throughput than \emph{FedBuff}. Ref. \cite{DBLP:conf/icpads/HaoZZ20} proposes a semi-asynchronous FL mechanism, wherein local models of unselected nodes are cached for several iterations before uploading to mitigate existing stragglers.

However, there is a common problem with both asynchronous and semi-asynchronous frameworks. This issue arises due to asynchronous aggregation, where models or gradients uploaded by some clients may lag several updates behind the current global model.
This inconsistency problem can lead to inaccurate global model updates or even learning failures.
To tackle this problem, we propose a novel time-driven synchronized FL framework.

\section{Time-driven Synchronized FL}\label{sec_sysmod}
In this section, we first present the proposed time-driven synchronized federated learning (T-SFL) framework. Afterwards, we evaluate its performance by analyzing the upper bound on the global loss function.

\subsection{Proposed T-SFL Framework}
\begin{figure}[t]
  \centering
  \includegraphics[width=1\linewidth]{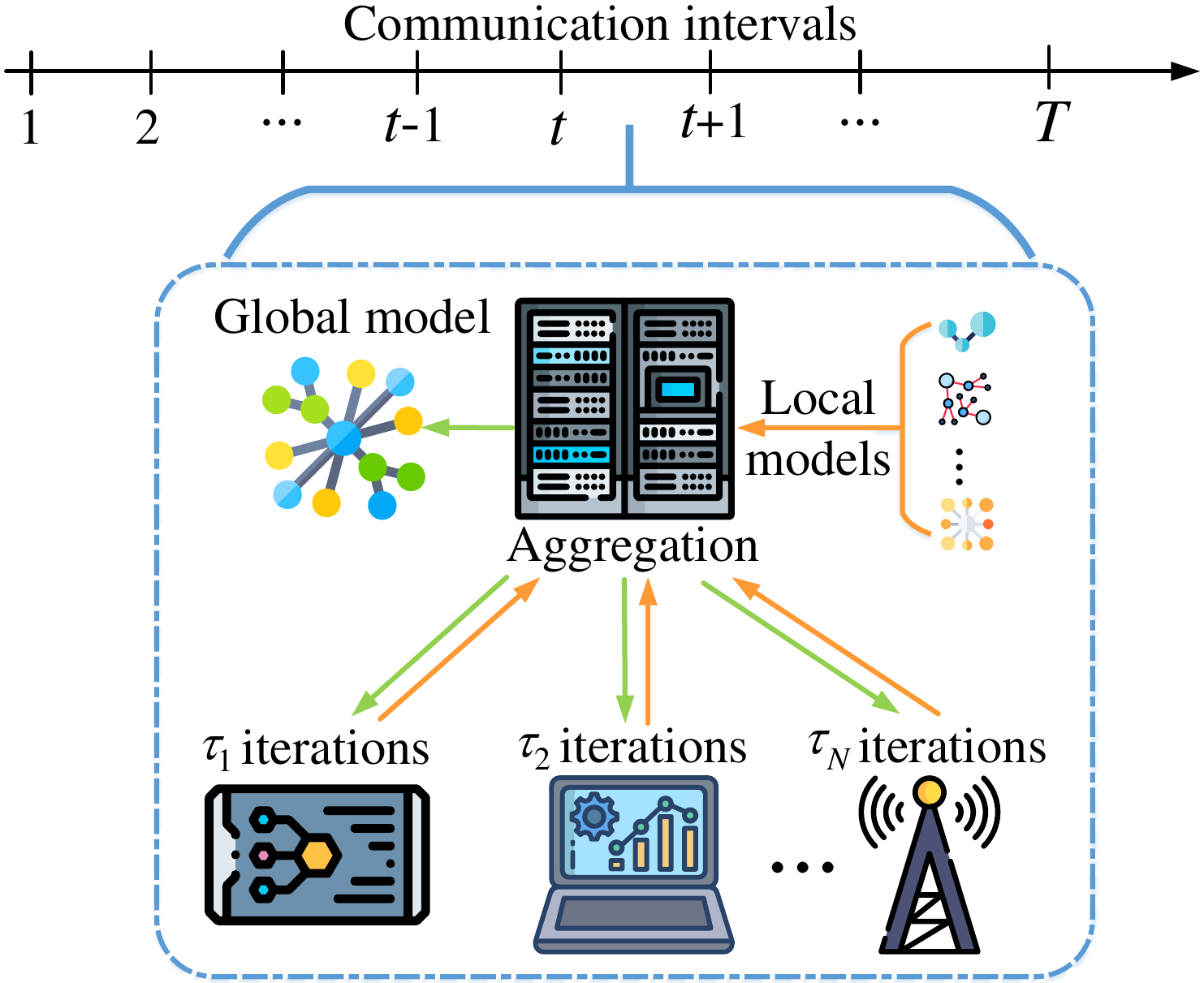}\\
  \caption{System model of T-SFL. For example, during the $t$-th communication interval, the server generates the global model based on the individual models submitted by each client with the varying numbers of iterations $\tau_1, \tau_2,...,\tau_N$. }\label{fig_system_model}
\end{figure}

The proposed T-SFL framework allows the global aggregation to be performed according to a predefined schedule in the time domain. The entire training process is divided into multiple communication time intervals, during which each client trains their local model instead of training a fixed number of iterations.
As shown in Fig.~\ref{fig_system_model}, the server aggregates the local models at regular intervals, referred to as communication intervals.
At the end of each interval, every client uploads their trained local model, and the server aggregates these uploaded models to generate the global model.
Due to the heterogeneity of the data size and computing power in the system, the number of local training iterations at each client varies from others.
We denote the number of local training iterations of the $i$-th client at the $t$-th communication interval as $\tau_i^t$.

Notably, at the end of each interval, every client uploads their locally trained model, which is then aggregated by the server.
This time-driven synchronization ensures that T-SFL's global model aggregation is both efficient in terms of reducing latency and capable of overcoming inconsistency in heterogeneous systems.
Simultaneously, clients within the T-SFL framework exhibit a high degree of autonomy. They possess the flexibility to join or withdraw from FL training at their discretion, allowing them to choose optimal times for participation within each communication interval. This autonomy has no impact on the global model aggregation or the overall system operation of T-SFL. Consequently, T-SFL proves to be an appealing choice, as it encourages increased client engagement in service computing applications.
In addition, T-SFL is well-suited for online learning scenarios due to its resilience against client drop-outs and intermittent participation.
Furthermore, T-SFL ensures predictability in training completion times by adhering to a predefined schedule, aligning seamlessly with the requirements of practical service computing applications.
Although T-SFL offers the aforementioned advantages, analyzing the performance of global aggregation in T-SFL introduces new challenges. This complexity arises from the variability in the number of training iterations among local models.

\subsection{Performance Analysis}

The upper bound on the loss function has been utilized to analyze the performance of FL in existing works, including \cite{9664296, 8664630, DBLP:conf/iclr/LiHYWZ20, DBLP:journals/corr/abs-2006-06954}.
To this end, we introduce the following definitions and assumptions.

\begin{definition}System parameters \cite{DBLP:conf/iclr/LiHYWZ20, DBLP:journals/corr/abs-2006-06954}
\begin{enumerate}
\item[1)] Aggregation weights $\{\rho_i^t\}$, i.e., $\boldsymbol{w}^{t}=\sum_{i=1}^N \rho^t_i \boldsymbol{w}^{t}_{i}$, $\sum_{i=1}^N \rho^t_i = 1$, and $F(\boldsymbol{w})=\sum_{i=1}^N \rho^t_i F_i(\boldsymbol{w})$.
\item[2)] Gradient $\bar{\boldsymbol{g}}^{t}_{i,h} = \nabla F_i(\boldsymbol{w}_{i,h}^t) = \mathbb{E}_{\xi_{i,h}^t} [\boldsymbol{g}^{t}_{i,h}]$, where $h$ denotes the $h$-th local iteration, $\xi_{i,h}^t$ denotes a mini-batch, $\bar{\boldsymbol{g}}^{t}_{i,h}$ and $\boldsymbol{g}^{t}_{i,h}$ represent the full batch gradient and the stochastic gradient, respectively.
\item[3)] Weight limit. Ref. \cite{DBLP:journals/corr/abs-2006-06954} has defined that $\rho^t_i \leq\frac{\theta}{N}, \forall i, t$, $\eta L (1+\theta)\geq1$, where $\eta$ denotes the learning rate and $\theta$ denotes the weight limit parameter, e.g., when $\theta=1$, we have $\rho^t_i=\frac{1}{N}, \forall i$.
\end{enumerate}
\end{definition}

\begin{definition}Auxiliary functions \cite{DBLP:journals/corr/abs-2006-06954}
\begin{enumerate}
\item[1)] Define a variable queue $\{\alpha^{t}_{i,h}\}, h \in \{0, ..., (H-1)\}$,  where $H$ denotes the maximum of local iterations among all clients. Here, $\alpha^{t}_{i,h} \in \{0,1\}$ denotes whether the $i$-th client participates in the $h$-th iteration ($0$ for negative while $1$ for positive), and $\sum^{H-1}_{h=0} \alpha^{t}_{i,h} = \tau^{t}_{i}$, where $\tau^{t}_{i}$ denotes the number of local iterations of the $i$-th client at the $t$-th interval.
\item[2)] Define an auxiliary function of the DML system, where
\begin{equation}
\bar{\boldsymbol{w}}^{0}=\boldsymbol{w}^{0}, \quad \bar{\boldsymbol{w}}^{tH+h+1}=\bar{\boldsymbol{w}}^{tH+h}-\eta \sum^{N}_{i=1} \rho^t_i \alpha^{t}_{i,h} \boldsymbol{g}^{t}_{i,h}. 
\end{equation}
\end{enumerate}
\end{definition}

Using the definitions and the stochastic gradient descent (SGD) algorithm in FL, we can derive that $\bar{\boldsymbol{w}}^{tH}=\boldsymbol{w}^{t}, \forall t$.
To facilitate the analysis of the convergence upper bounds, we then introduce the following assumptions, which are commonly used in convergence bound analysis \cite{DBLP:conf/iclr/LiHYWZ20, DBLP:journals/corr/abs-2006-06954, DBLP:conf/aaai/YuYZ19}.
\begin{assumption}Suppose that, for $\forall i$,
\begin{enumerate}
\item[1)] $\mathbb{E}[\Vert \boldsymbol{g}^{t}_{i,h} \Vert_2^2]\leq G^2$, and
$\mathbb{E}[\Vert \boldsymbol{g}^{t}_{i,h}-\bar{\boldsymbol{g}}^{t}_{i,h} \Vert_2^2]\leq \sigma_i^2$.
\item[2)] $F_i$ is $L$-smooth, i.e., $\forall \boldsymbol{v}$ and $\forall \boldsymbol{w}$, $F_i(\boldsymbol{v}) \leq F_i(\boldsymbol{w}) + (\boldsymbol{v} - \boldsymbol{w})^T \nabla F_i(\boldsymbol{w}) + \frac{L}{2}\Vert \boldsymbol{v} - \boldsymbol{w} \Vert_2^2$.
\item[3)] $F_i$ is $L$-smooth in expectation, i.e., $\forall \boldsymbol{w}$, $\mathbb{E}[\Vert \nabla F_i(\boldsymbol{w}) - \nabla F_i(\boldsymbol{w}_i^*) \Vert_2^2 ] \leq 2L ( F_i(\boldsymbol{w}) - F_i(\boldsymbol{w}_i^*) )$.
\item[4)] $F_i$ is non-convex.
\end{enumerate}
\end{assumption}

Using the above definitions and assumptions, we can derive the upper bound on the loss function of the T-SFL framework as stated in \textbf{Theorem 1}.

\begin{theorem}{The upper bound on the loss function of T-SFL is expressed as}\label{threm_final}
\begin{align}\label{eq_4}
\frac{1}{L^2} \Vert \nabla F(\boldsymbol{w}^T)\Vert^2 \leq   \frac{ \Vert \boldsymbol{w}^0 - \boldsymbol{w}^* \Vert^2 +X +Y + Z} {W},
\end{align}
where
\begin{align}\label{eq_5}
\quad &X= \eta^3 L (H-1) G^2 \sum_{i=1}^N \sum_{t=0}^{T-1} \rho^t_i(\tau_i^t)^2 , \nonumber  \\
&Y= \eta^2 N  \sum_{i=1}^N \sigma_i^2 \sum_{t=0}^{T-1} (\rho^t_i)^2 \tau_i^t, \nonumber \\
&Z= (2\eta L (\eta L (1+\theta) - 1) +\eta^2) \sum_{i=1}^N \sum_{t=0}^{T-1} \rho^t_i \tau_i^t \Gamma_i , \nonumber \\
&W = 1 + 2 \eta L [1 - \eta L (1+\theta) ] \sum_{i=1}^N \sum_{t=0}^{T-1} \rho^t_i \tau_i^t,
\end{align}
$\Gamma_i = \Vert \boldsymbol{w}^* - \boldsymbol{w}_i^* \Vert^2$, $\boldsymbol{w}^*$ denotes the optimal global model, $\boldsymbol{w}_i^*$ denotes the optimal local model, $\frac{\Vert \boldsymbol{w}^0 - \boldsymbol{w}^* \Vert_2^2}{W}$ represents the initial item, $\frac{X}{W}$ represents the gradient item, $\frac{Y}{W}$ represents the mini-batch item, and $\frac{Z}{W}$ represents the data distribution item, i.e., not identically and independently distributed (non-IID) degree, respectively.
\end{theorem}

\begin{IEEEproof}
Please refer to Appendix \ref{apd_B} for more details.
\end{IEEEproof}

From \textbf{Theorem \ref{threm_final}}, we can obtain that the set of aggregation weights $\{\rho_i^t\}$ can yield different upper bound values. 
In this context, we propose a robust aggregation algorithm to minimize the upper bound, which can enhance the learning accuracy and robustness of T-SFL in heterogeneous systems.

\section{Discriminative Model Select Algorithm}\label{sec_alg}
In this section, we first optimize the aggregation weights to minimize the loss function of T-SFL, considering different numbers of local iterations $\tau_i^t$ at each client. Then, we propose a robust DMS algorithm to enhance the fairness, security and learning performance. Finally, we analyze the convergence performance of the DMS algorithm.

\subsection{Optimal Aggregation Weights}
Building upon \textbf{Theorem \ref{threm_final}}, we take the first derivative of $\rho_i^t$ and present the following theorem.

\begin{theorem}{The optimal aggregation weights that minimize the upper bound on loss function can be expressed as}\label{theo_rho}
\begin{align}\label{eq_rho}
\rho_i^* = \frac{B D + ( A \Gamma_i  + C \tau_i^t ) (1 + B \sum_{j=1}^N \sum_{t=0}^{T-1} \rho_j^* \tau_j^t) }{(1 + B \sum_{j=1}^N \sum_{t=0}^{T-1} \rho_j^* \tau_j^t) \cdot (2 \eta^2 N \sigma_i^2) },
\end{align}
\begin{align}\label{eq_rho1}
A= &\eta \{2 L[\eta L (1+\theta) - 1 ] + \eta \} , \nonumber  \\
B = &2 \eta L [1 - \eta L (1+ \theta)], \nonumber \\
C = &\eta^3 L (H-1) G^2, \notag \\
D = &\Vert \boldsymbol{w}^0 - \boldsymbol{w}^* \Vert^2 +\eta^2 N \sigma_i^2 \sum_{j=1}^N \sum_{t=0}^{T-1} (\rho_j^*)^2 \tau_j^t \notag \\
&+ A \Gamma_i \sum_{j=1}^N \sum_{t=0}^{T-1} \rho_j^* \tau_j^t + C \sum_{j=1}^N \sum_{t=0}^{T-1} \rho_j^* (\tau_j^t)^2,
\end{align}
where $i \in \{ 1,2,...,N\}$.
\end{theorem}
\begin{IEEEproof}
Please refer to Appendix \ref{apd_rho} for more details.
\end{IEEEproof}

Based on \textbf{Theorem \ref{theo_rho}}, we observe that given $\sigma_i$ and $\Gamma_i$, the optimal aggregation weight $\rho_i^*$ increases as the number of local iterations $\tau_i^t$ increases.
This suggests that clients who perform more local iterations are allocated with a higher weight.
It aligns with intuition since clients who perform more iterations are more likely to produce a high-quality training model \cite{9705108}.

However, deriving the optimal weights from \textbf{Theorem \ref{theo_rho}} is quite challenging since it involves multiple variables, including $\sum_{i=1}^N \sum_{t=0}^{T-1} (\rho_i^*)^2 \tau_i^t$, $\sum_{i=1}^N \sum_{t=0}^{T-1} \rho_i^* \tau_i^t$ and $\sum_{i=1}^N \sum_{t=0}^{T-1} \rho_i^* (\tau_i^t)^2$.
To overcome this challenge and improve both the fairness and security, we propose a robust aggregation algorithm in the following subsection.

\subsection{Proposed Discriminative Model Select Algorithm}
First, we obtain the following corollary from \textbf{Theorem \ref{theo_rho}}.
\begin{corollary}{Given $\sigma_i = \sigma$ and $\Gamma_i = \Gamma, \forall i$, we have} \label{coro_1}
\begin{align}\label{eq_8}
\rho_i^* - \rho_j^* = \frac{\eta L (H-1) G^2 }{ 2  N \sigma^2 }  (\tau_i^t - \tau_j^t),
\end{align}
\end{corollary}
where $i \in \{ 1,2,...,N\}$ and $j \in \{ 1,2,...,N\}$.
\begin{IEEEproof}
The proof is straightforward from (\ref{eq_rho}) and (\ref{eq_rho1}).
\end{IEEEproof}

Notably, this paper primarily addresses the challenge of heterogeneous iteration times among local models, arising from variations in computing resources and dataset sizes among clients. To streamline the analysis and derive a straightforward optimal weight allocation formula in the presence of system heterogeneity, we assume IID data when deducing \textbf{Corollary \ref{coro_1}}. This assumption is made to eliminate the influence of data distribution, ensuring a more concise and clear formulation as stated in (\ref{eq_8}).
It is noteworthy that in FL, the non-IID problem pertains to data heterogeneity, distinct from the emphasis of this article on system resource heterogeneity.
While considering non-IID settings, the assignment of optimal weights may deviate from the findings in \textbf{Corollary \ref{coro_1}}, which is extensively explored in the existing literature \cite{DBLP:journals/corr/abs-2006-06954, 10.5555/3495724.3496362}.
Fortunately, our experimental results demonstrate that the aggregation weights derived from \textbf{Corollary \ref{coro_1}} yield considerable accuracy improvements even in non-IID settings, as shown in Sec. \ref{sec_exp}.

From (\ref{eq_8}), we can see that the difference between any two optimal aggregation weights is a linear function of the difference between the corresponding number of iterations.
Therefore, we can simplify the calculation in \textbf{Theorem \ref{theo_rho}} using \textbf{Corollary \ref{coro_1}}. Afterwards, by applying the normalization constraint $\sum_{i=1}^N \rho_i^t=1$, we can set up $N$-ary linear equations to obtain the solution set of optimal weights.

\begin{algorithm}[t]
	\caption{Discriminative Model Select Algorithm} \label{alg_WF}
	\LinesNumbered
\KwData{$\eta$, $T$, $N$, $G$, $\sigma$ and $H$}
         {Collect $\tau_i^t$ from all clients at the $t$-th communication interval and update $H= \max \limits_i\{\tau_i^t\}$}\\
Calculate threshold $K^t$, filtering probability $\mathcal{P}_i^t$, and $\beta_i^t$ as (\ref{eq_wl1}), (\ref{eq_p}), and (\ref{eq_threshold}), respectively\\
Assign zero weight to the $i$-th client if $\beta_i^t = 0$ \\
Obtain the set of optimal weights $\{\rho_i^*\}$ by (\ref{eq_8}), (\ref{eq_cf}), and $\sum_{i=1}^N \rho_i^t = 1$\\
        \KwResult{ $\{\rho_i^*\}$}
\end{algorithm}

However, in T-SFL, the number of local training iterations among clients may vary significantly even within the same interval.
This means that the optimal aggregation weights obtained from (\ref{eq_8}) may not be fair since they are also allocated to clients with insufficient training iterations.
This issue can adversely affect the performance of the global model and reduce the learning accuracy and convergence rate.
To address this problem, we propose a discriminative model selection (DMS) algorithm to optimize the set of aggregation weights in each communication interval.
The DMS algorithm is designed to promote fairness, efficiency, robustness, and security of T-SFL, as shown in \textbf{Algorithm \ref{alg_WF}}.

First, we design the iteration threshold of the DMS algorithm at the $t$-th communication interval as
\begin{equation}\label{eq_wl1}
  K^t = \frac{1}{N} \sum_{i=1}^N \tau_i^t.
\end{equation}
If the $i$-th client's iteration number $\tau_i^t$ is lower than the threshold, there is a filtering probability $\mathcal{P}_i^t$ that the DMS algorithm assigns it a zero aggregation weight. We can denote the filtering probability $\mathcal{P}_i^t$ as
\begin{equation}\label{eq_p}
  \mathcal{P}_i^t = \frac{K^t - \tau_i^t}{H} ,\quad \mathrm{if} \quad \tau_i^t < K^t.
\end{equation}

From (\ref{eq_p}), we notice that the filtering probability $\mathcal{P}_i^t$ increases as the difference between $K^t$ and $\tau_i^t$ increases.
This means that clients with fewer iterations are more likely to be assigned a zero aggregation weight, promoting fairness in the system.
It is in line with the intuition that clients with a smaller number of iterations contribute less to the learning performance of T-SFL, and thus, should be filtered out of the aggregation with a higher probability.
We can then define a Bernoulli distribution variable $\beta_i^t$ to indicate whether the $i$-th client is assigned an aggregation weight at the $t$-th interval. Note that $\beta_i^t \in \{0,1\}$ , where
\begin{equation}\label{eq_threshold}
  \beta_i^t = \left\{
  \begin{aligned}
    &0, & \mathrm{with} \quad \mathrm{probability} \quad \mathcal{P}_i^t , \quad\\
    &1,  & \mathrm{with} \quad \mathrm{probability} \quad 1-\mathcal{P}_i^t,
  \end{aligned}
  \right.  \mathrm{if} \quad \tau_i^t < K^t.
\end{equation}

Then, we can state the following theorem.

\begin{theorem}{Under the DMS algorithm, the optimal aggregation weights are given by }\label{theo_cf}
\begin{align}\label{eq_cf}
  \rho_i^* = \left\{
  \begin{aligned}
    &\frac{B E + F ( A \Gamma_i   + C \tau_i^t  )  }{F  (2 \eta^2 N \sigma_i^2 ) }, & \mathrm{if} \quad \beta_i^t = 1,\\
    & \qquad \qquad \quad 0,  &  \mathrm{if} \quad \beta_i^t = 0,
  \end{aligned}
  \right.
\end{align}
\begin{align}\label{eq_cf1}
E = &\Vert \boldsymbol{w}^0 - \boldsymbol{w}^* \Vert^2 +\eta^2 N \sigma_i^2 \sum_{j=1}^N \sum_{t=0}^{T-1} (\rho_j^*)^2 \tau_j^t \beta_i^t \notag \\
&+ A \Gamma_i \sum_{j=1}^N \sum_{t=0}^{T-1} \rho_j^* \tau_j^t \beta_i^t + C \sum_{j=1}^N \sum_{t=0}^{T-1} \rho_j^* (\tau_j^t)^2 \beta_i^t, \notag \\
F = &1 + B \sum_{j=1}^N \sum_{t=0}^{T-1} \rho_j^* \tau_j^t \beta_i^t,
\end{align}
where $i \in\{1,2,...,N\}$.
\end{theorem}
\begin{IEEEproof}
The proof is straightforward and involves substituting the variable $\beta_i^t$ into the proof of \textbf{Theorem \ref{theo_rho}}.
\end{IEEEproof}

Using (\ref{eq_8}) and (\ref{eq_cf}), we can obtain the optimal weight set $\{\rho_i^*\}$.
It is noteworthy that clients with a lower number of iterations are more inclined to exhibit $\beta_i^t=0$, as observed from (\ref{eq_p}) and (\ref{eq_threshold}).
Therefore, we assign a weight of zero to clients with $\beta_i^t = 0$ in \emph{line 3}.
Finally, we calculate the optimal aggregation weight set $\{ \rho_i^* \}$ by (\ref{eq_8}) and (\ref{eq_cf}) in \emph{line 4}.

The DMS algorithm can effectively promote fairness and enhance the quality of aggregated models, leading to faster convergence of FL.
The iteration detection feature of the DMS algorithm allows for the mitigation of the negative effects of clients with low computing power while preventing malicious clients from engaging in free-riding behavior.
This, in turn, improves the final learning accuracy, robustness, and security of T-SFL.
Therefore, using the DMS algorithm, the server can aggregate high-quality models to generate a global model that has the lowest loss function at the $t$-th communication interval. 

\subsection{Convergence Analysis}
Let us introduce some definitions and assumptions that are commonly used in convergence analysis of FL \cite{DBLP:conf/mlsys/LiSZSTS20, DBLP:conf/bigdataconf/ChenNSR20}.

\begin{definition}Bounded gradient dissimilarity\label{defin_2}
\begin{enumerate}
\item[1)]  The local functions $F_i(\cdot)$ are $V$-locally dissimilar at $\boldsymbol{w}$ if $\mathbb{E}_i [\Vert \nabla F_i (\boldsymbol{w}) \Vert^2] \leq V^2 \Vert \nabla F (\boldsymbol{w}) \Vert^2, \forall i$.
\end{enumerate}
\end{definition}

\begin{assumption}\label{assumption_nonconvex} Suppose that, $\forall i$,
\begin{enumerate}
\item[1)] The global loss function $F (\boldsymbol{w})$ is bounded, i.e., $F_{\mathrm{min}} = F(\boldsymbol{w}^*) > - \infty$.

\item[2)] There exists $\epsilon >0$ such that $\nabla F (\boldsymbol{w})^\top \mathbb{E}_i [\nabla F_i (\boldsymbol{w})] \geq \epsilon \Vert \nabla F (\boldsymbol{w}) \Vert^2$ holds for all $\boldsymbol{w}$. Note that if $\epsilon = 1$, then $\nabla F_i (\boldsymbol{w})$ is an unbiased estimator of $\nabla F (\boldsymbol{w})$.
\end{enumerate}
\end{assumption}

Based on the definitions and assumptions, we can derive the convergence bound of gradients for the DMS algorithm.
\begin{theorem}\label{theorem_5}
Under DMS, the mean value of cumulative gradients on the global loss function is upper-bounded by
  \begin{equation}\label{eq_10}
    \frac{1}{T} \sum_{t=0}^{T-1} \Vert \nabla F(\boldsymbol{w}^t) \Vert^2 \leq   \frac{2 \eta}{T (2 \epsilon - \eta L   V^2)}   [F(\boldsymbol{w}^0) - F(\boldsymbol{w}^*)],
  \end{equation}
if $\eta<\frac{2 \epsilon}{V^2   L}$.
\end{theorem}

\begin{IEEEproof}
Please refer to Appendix \ref{app_E} for more details.
\end{IEEEproof}

According to (\ref{eq_10}), the mean value of cumulative gradients decreases and approaches zero as the total number of communication intervals $T$ increases. In this context, it can be concluded that the DMS algorithm converges as $T$ becomes sufficiently large.

\section{Experimental Results}\label{sec_exp}
In this section, we first compare the test accuracy and training time of conventional SFL, AFL, and T-SFL.
Following that, we then evaluate the learning accuracy, convergence rate, and client participation probability of the T-SFL system using the DMS algorithm.
\subsection{Experimental Setting}

1) In our experiments, we use four datasets under the non-IID setting to showcase the learning accuracy, training time, and convergence rate of federated learning in heterogeneous systems. The datasets are presented below.

$\bullet$ MNIST. The standard MNIST handwritten digit recognition dataset comprises 60,000 training and 10,000 testing examples. 
Each example is a grayscale image of a handwritten digit with 28$\times$28 pixels, ranging from 0 to 9. 

$\bullet$ Cifar-10. Cifar-10 is a compact dataset curated by Alex Krizhevsky and Ilya Sutskever, students of Hinton, to recognize universal objects. It contains 10 categories of RGB color images, namely airplane, automobile, bird, cat, deer, dog, frog, horse, ship, and truck. Each image has a size of 32$\times$32 pixels, and each category comprises 6000 images. It consists of 50,000 training images and 10,000 test images.

$\bullet$ Fashion-MNIST. The Fashion-MINST dataset consists of 10 distinct clothing categories, including T-shirts, trousers, pullovers, dresses, coats, sandals, shirts, sneakers, bags, and ankle boots. This dataset is more intricate than MNIST and thus provides a more accurate evaluation of the neural networks' performance in practical applications.

$\bullet$ SVHN. The Street View House Number (SVHN) dataset is derived from Google street view house numbers and comprises over 600,000 digit images. Unlike MNIST, SVHN is a significantly more challenging, unsolved real-world problem that involves recognizing digits and numbers in natural scene images. With an order of magnitude more labeled data than MNIST, SVHN represents a more realistic and complex dataset for evaluating the learning performance of neural networks.

2) T-SFL setting. As a default configuration, we initialize the number of clients $N = 20$, with a dataset size of $\vert \mathcal{D}_i \vert = 1024$ and a batch size of $\vert \xi_i \vert = 32$ for all clients. The learning rate is set to $\eta= 0.003$ for MNIST and Fashion-MNIST (F-MNIST), $\eta=0.005$ for Cifar-10 and SVHN. Additionally, we set the total number of communication intervals $T = 50$ for MNIST and Cifar-10, $T = 100$ for F-MNIST and SVHN.

3) Static cases. We examine two computing power cases in a static setting, with the data size set to default. In Case 1, half of the clients have a local training iteration number of $\tau=1$, while the other half have a number of $\tau=4$. In Case 2, a quarter of the clients have a local training iteration number of $\tau=1$, another quarter have a number of $\tau=2$, a third quarter have a number of $\tau=3$, and the remaining clients have a number of $\tau=4$.

4) Dynamic cases. We set up a dynamic setting where each client has a time-varying computing power and data size case. In Case 3, the computing power ratio distribution is as follows: a quarter of the clients have $\tau_i^t$ values drawn from $\lfloor \mathcal{N}(2,0.4) \rfloor$, another quarter have $\tau_i^t$ values drawn from $\lfloor \mathcal{N}(3,0.6) \rfloor$, a third quarter have $\tau_i^t$ values drawn from $\lfloor \mathcal{N}(4,0.8) \rfloor$, and the remaining clients have $\tau_i^t$ values drawn from $\lfloor \mathcal{N}(5,1) \rfloor$, where $\lfloor \cdot \rfloor$ denotes the floor function and $\mathcal{N}(\cdot)$ denotes the Gaussian distribution. Additionally, the data size distribution is as follows: one fifth of the clients have $\vert \mathcal{D}_i \vert$ values drawn from $\lfloor \mathcal{N}(512, 100) \rfloor$, another fifth have $\vert \mathcal{D}_i \vert$ values drawn from $\lfloor \mathcal{N}(768, 150) \rfloor$, a third fifth have $\vert \mathcal{D}_i \vert$ values drawn from $\lfloor \mathcal{N}(1024, 200) \rfloor$, a fourth fifth have $\vert \mathcal{D}_i \vert$ values drawn from $\lfloor \mathcal{N}(1280, 250) \rfloor$, and the final fifth have $\vert \mathcal{D}_i \vert$ values drawn from $\lfloor \mathcal{N}(1536, 300) \rfloor$.

5) Learning models. We utilize the cross-entropy function as the loss function. For the MNIST dataset, we implement a convolutional neural network (CNN) consisting of two convolutional layers. The first layer has a 5$\times$5 kernel with 10 channels, followed by ReLU activation and a 2$\times$2 max-pooling layer. The second layer also has a 5$\times$5 kernel, but with 20 channels, followed by ReLU activation and another 2$\times$2 max-pooling layer. This is then followed by two fully connected layers (320$\times$10).
For the Cifar-10 dataset, we employ the well-known VGG11 network.
For the Fashion-MNIST dataset, we use a CNN comprising six 3-kernel convolutional layers, each followed by a ReLU layer and a 2$\times$2 pooling layer.
To handle the SVHN dataset, we employ the widely recognized ResNet network.

6) Hardware and software. We conduct all experiments using PyTorch on 20 personal computers equipped with an NVIDIA GeForce RTX 3070 GPU and an 11th Gen Intel(R) Core(TM) i7-11800H @2.30GHZ CPU. Additionally, we utilize a server with the same configuration. The personal computers communicate with the server through a wired network. The reported experimental results represent the average of 10 independent experiments.
\subsection{Performance Under Different System Heterogeneity}
\begin{figure}[t]
  \centering
  \includegraphics[width=1\linewidth]{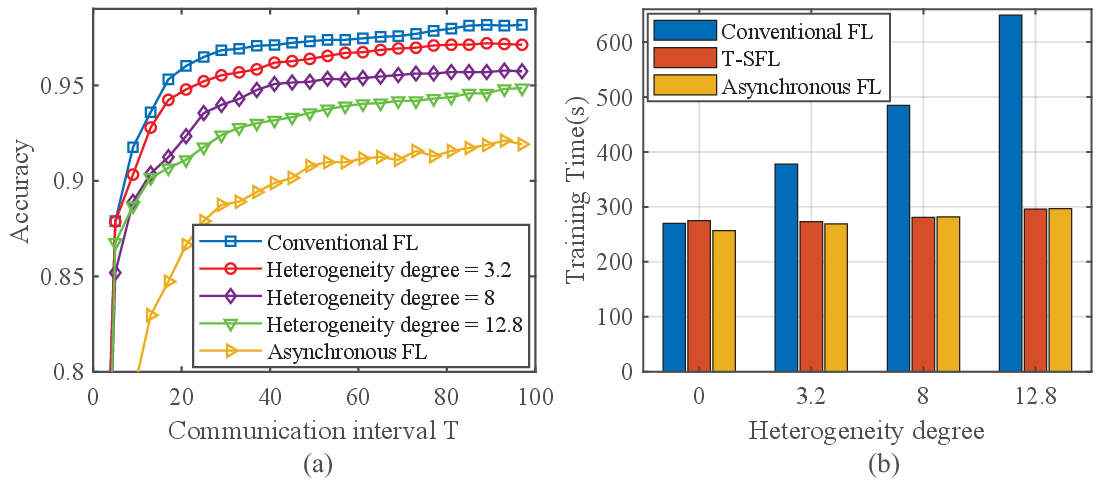}\\
  \caption{Performance comparison between conventional FL, AFL, and T-SFL. (a) Test accuracy of T-SFL versus $T$ on MNIST dataset for systems with varying degrees of heterogeneity. (b) Training time comparison on MNIST dataset for systems with varying degrees of heterogeneity.}\label{fig_heterogeneity}
\end{figure}
Fig. \ref{fig_heterogeneity}(a) plots the test accuracy of T-SFL on MNIST for a system with different heterogeneity degrees. Heterogeneity degree\footnote{As stated in \cite{DBLP:conf/icml/ZhuHDZ22}, the degree of heterogeneity is defined as $\delta = \frac{1}{N} \sum_{i=1}^N (\tau_i - \bar{\tau})^2$, where $\bar{\tau}$ represents the average local training iteration number across all clients.} refers to the degree of variation in the computing power distribution across different clients participating in the FL process.
The figure shows how the test accuracy of T-SFL changes as the heterogeneity degree varies.
Fig. \ref{fig_heterogeneity}(b) plots the training time of conventional SFL, T-SFL, and asynchronous FL on MNIST for a system with different heterogeneity degrees.
The figure shows how the training time of each framework changes as the heterogeneity degree varies.
First, it is observed that the learning accuracy of T-SFL decreases with an increase in system heterogeneity. This decrease in performance can be attributed to the presence of heterogeneous systems, where each client has varying computing power and data size.
Second, as the degree of heterogeneity decreases, the performance of T-SFL approaches that of conventional FL. On the other hand, as the degree of heterogeneity increases, T-SFL's performance becomes more similar to that of AFL.
The reason for this is that T-SFL and conventional FL are equivalent in a homogeneous system, where each client has the same amount of computing power and data size. However, in a highly heterogeneous system, it is possible that only one client can complete training during each communication interval, leading T-SFL to behave more like AFL.
Third, as the degree of heterogeneity increases, the training time of conventional FL also increases, while the training times of T-SFL and asynchronous FL remain relatively stable. This is because conventional FL experiences additional latency due to the varying computing power and data sizes of the clients, which is not the case for T-SFL and asynchronous FL.

\begin{figure}[t]
  \centering
  \includegraphics[width=1\linewidth]{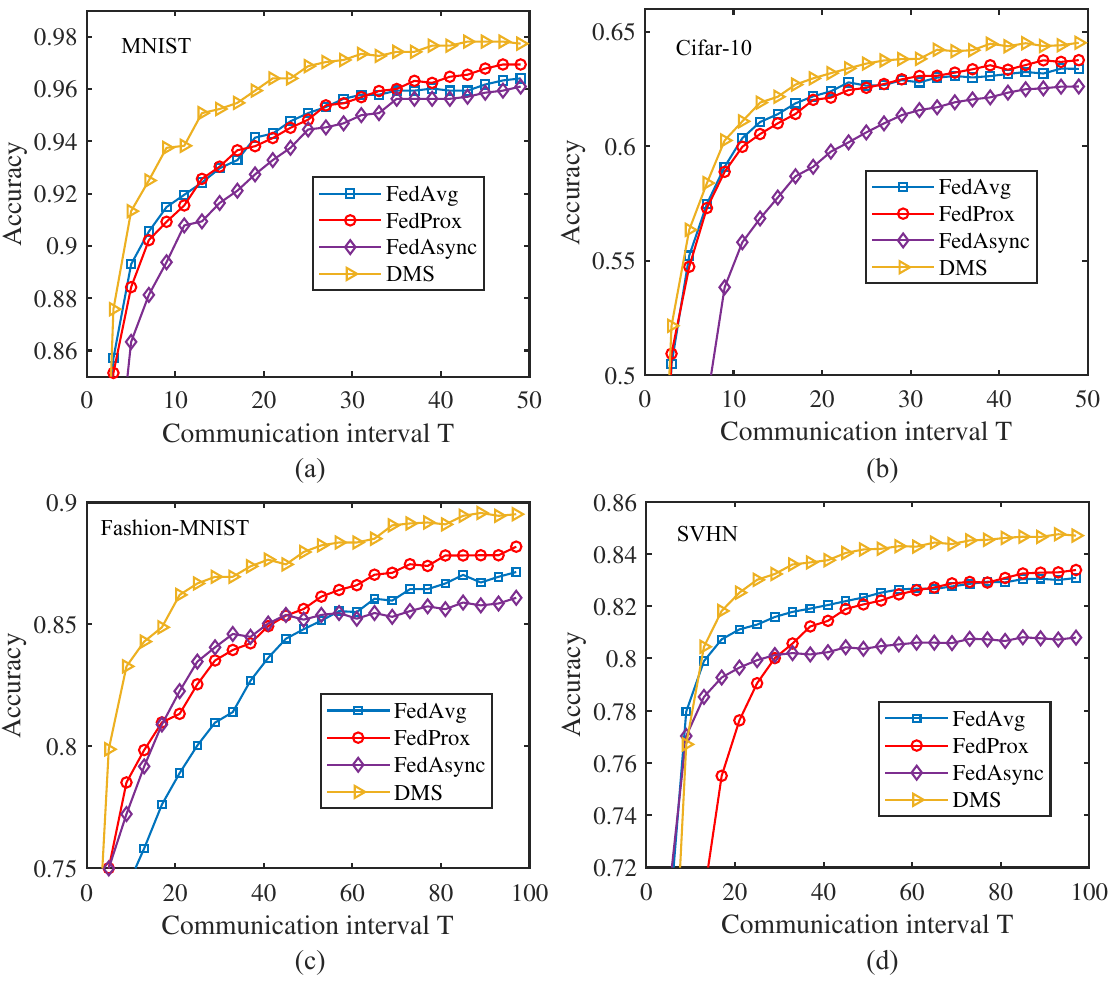}\\
  \caption{Test accuracy versus $T$ for (a) MNIST dataset, (b) Cifar-10 dataset, (c) F-MNIST dataset, and (d) SVHN dataset under Case 1. }\label{fig_case1}
\end{figure}
\begin{figure}[t]
  \centering
  \includegraphics[width=1\linewidth]{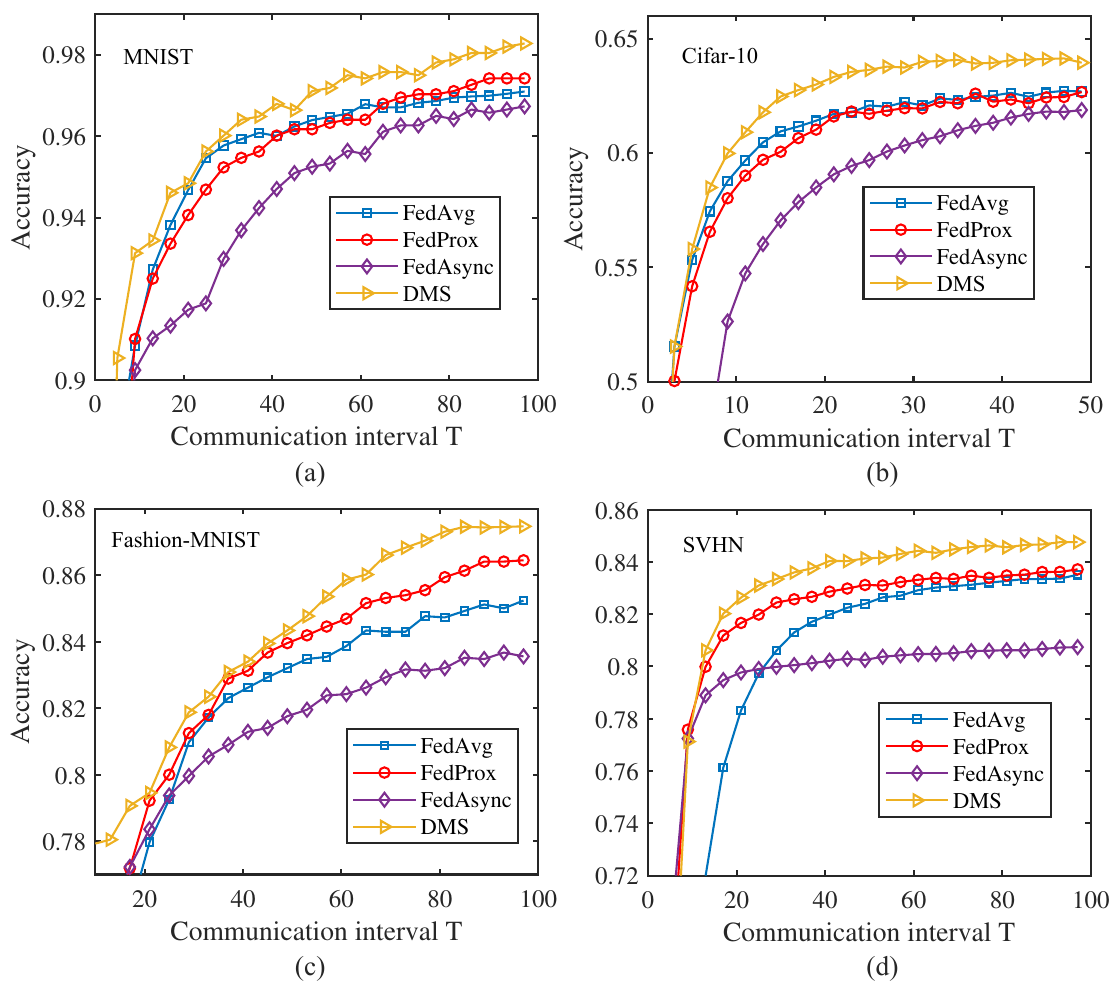}\\
  \caption{Test accuracy versus $T$ for (a) MNIST dataset, (b) Cifar-10 dataset, (c) F-MNIST dataset, and (d) SVHN dataset under Case 2. }\label{fig_case2}
\end{figure}

\begin{table}[t]
\centering
\caption{The highest global model accuracy on the MNIST, Cifar-10, Fashion-MNIST, and SVHN datasets in the static cases.}
\begin{tabular}{cccccc}
\Xhline{1.2pt}
& Dataset & \emph{FedAvg} & \emph{FedProx} &\emph{FedAsync} &  DMS  \\
\hline
\multirow{4}{*}{Case 1} & MNIST &96.41\% & 97.01\% & 96.17\% & \textbf{97.89\%}   \\
& Cifar-10 &63.54\% & 63.70\% & 62.68\% &  \textbf{64.57\%}  \\
& F-MNIST &87.15\% & 88.19\% & 86.09\% &  \textbf{89.52\%}  \\
& SVHN &83.08\% & 83.38\% & 80.80\% &  \textbf{84.70\%}  \\
\hline
\multirow{4}{*}{Case 2} & MNIST &97.18\% & 97.58\% & 96.73\% & \textbf{98.20\%}   \\
& Cifar-10 &62.70\% & 62.27\% & 61.89\% &  \textbf{64.11\%}  \\
& F-MNIST &85.23\% & 86.45\% & 83.55\% &  \textbf{87.47\%}  \\
& SVHN &83.51\% & 83.71\% & 80.74\% &  \textbf{84.76\%}  \\
\Xhline{1.2pt}
\end{tabular}
\label{tab_static}
\end{table}
\subsection{Performance of DMS Under Static Cases}
Fig.~\ref{fig_case1} plots the experimental results of the test accuracy on MNIST, Cifar-10, F-MNIST, and SVHN for various algorithms under Case 1.
Fig.~\ref{fig_case2} presents the experimental results of test accuracy on the same datasets under Case 2.
Table \ref{tab_static} lists the corresponding highest test accuracy of the global models.
First, it is observed that our proposed DMS algorithm outperforms other state-of-the-art algorithms in both cases with an average gain of 2\% in accuracy. This is particularly evident in Case 1, which has the highest degree of heterogeneity.
Second, it is shown that the performance gain achieved by the DMS algorithm decreases as the degree of heterogeneity decreases.
For instance, on the MNIST dataset, the DMS algorithm achieves a 1.72\% accuracy gain with a heterogeneity degree of 2.25 in Case 1, compared to a 1.47\% accuracy gain with a heterogeneity degree of 1.25 in Case 2.
This suggests that the DMS algorithm can effectively mitigate the performance loss caused by a heterogeneous system, and its mitigation effect is more significant in a system with a large degree of heterogeneity.
Third, an important finding from the experimental results is that the DMS algorithm demonstrates a comparable convergence rate to that of \emph{FedAvg} and \emph{FedProx} algorithms, but outperforms the \emph{FedAsync} algorithm in terms of convergence speed for all datasets and cases. This finding suggests that the DMS algorithm can strike a good balance between learning accuracy and convergence speed, making it a promising FL algorithm for practical applications.

\begin{figure}[t]
  \centering
  \includegraphics[width=1\linewidth]{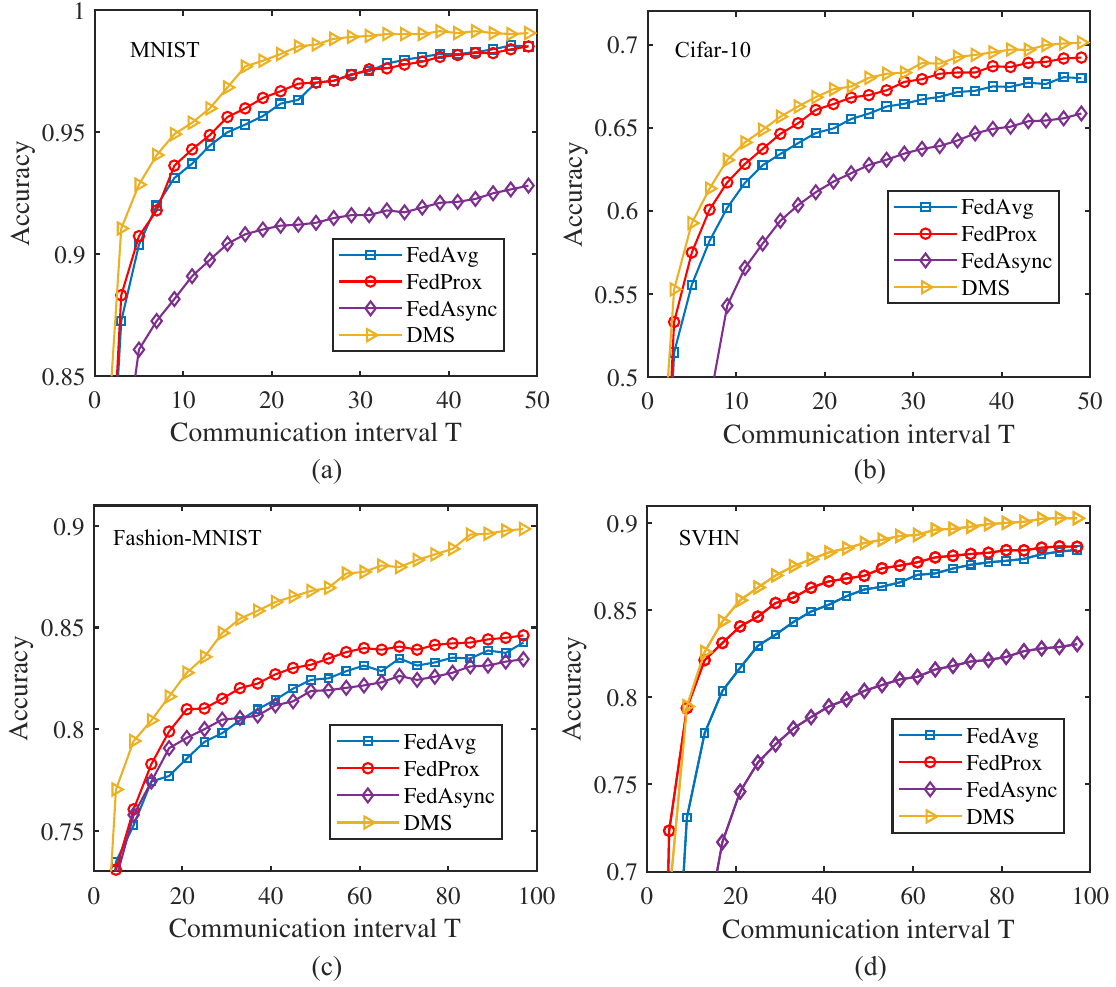}\\
  \caption{Test accuracy versus $T$ for (a) MNIST dataset, (b) Cifar-10 dataset, (c) F-MNIST dataset, and (d) SVHN dataset under Case 3.  }\label{fig_dynamic} 
\end{figure}

\begin{table}[t]
\centering
\caption{The highest global model accuracy on the MNIST, Cifar-10, Fashion-MNIST, and SVHN datasets in the dynamic cases.}
\begin{tabular}{cccccc}
\Xhline{1.2pt}
& Dataset & \emph{FedAvg} & \emph{FedProx} &\emph{FedAsync} &  DMS  \\
\hline
\multirow{4}{*}{\makecell[c]{Case 3}}   &MNIST & 98.67\% & 98.55\% & 92.97\% &  \textbf{99.06\%}   \\
& Cifar-10 &68.07\% & 69.21\% & 65.97\% &  \textbf{70.29\%}  \\
& F-MNIST &84.26\% & 84.61\% & 83.44\% &  \textbf{89.84\%}  \\
& SVHN & 88.46\% &  88.64\% &  83.05\% &  \textbf{ 90.27\%}  \\
\hline
$2$-iteration& MNIST &98.13\% & 99.14\% & 97.75\% &  \textbf{99.57\%}  \\
\hline
$4$-iteration& MNIST &97.11\% & 97.46\% & 96.58\% &  \textbf{98.59\%}  \\
\Xhline{1.2pt}
\end{tabular}
\label{tab_dynamic}
\end{table}

\begin{figure}[t]
  \centering
  \includegraphics[width=0.8\linewidth]{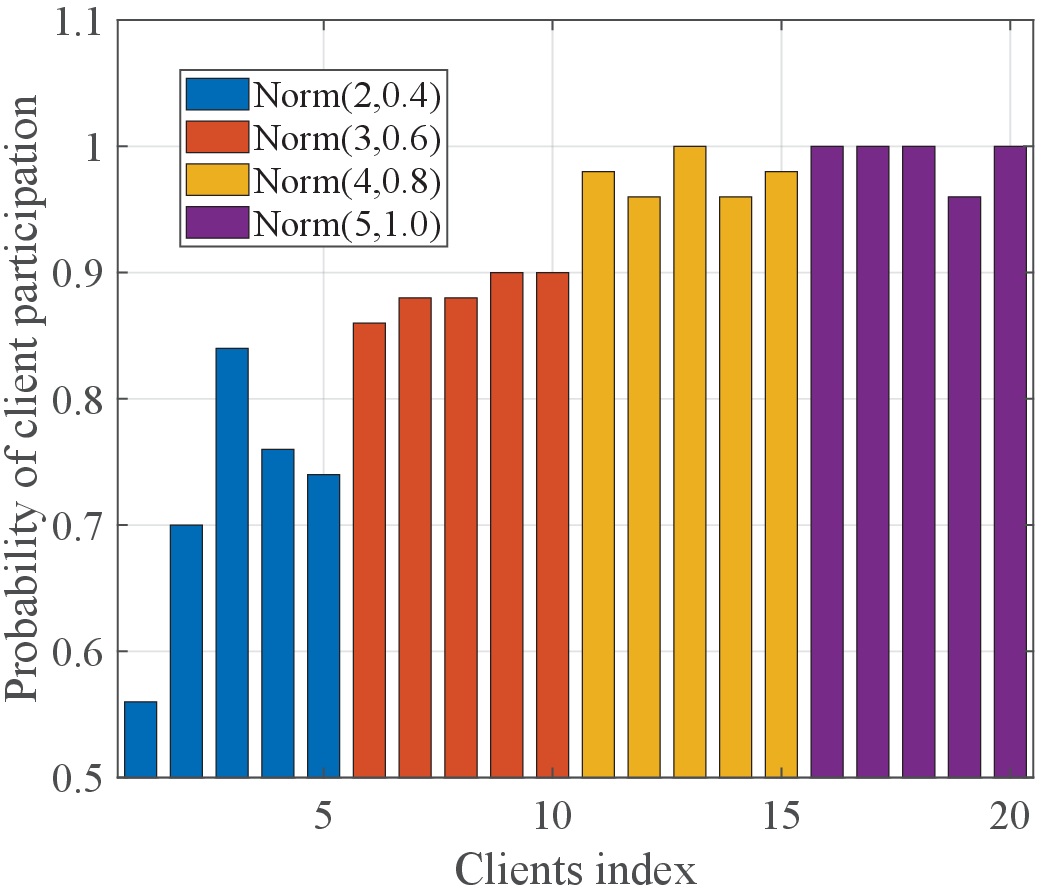}\\
  \caption{The probabilities of client participation for T-SFL under Case 3, where the computing power follows the Normal distribution. }\label{fig_bar}
\end{figure}


\begin{figure}[t]
  \centering
  \includegraphics[width=1\linewidth]{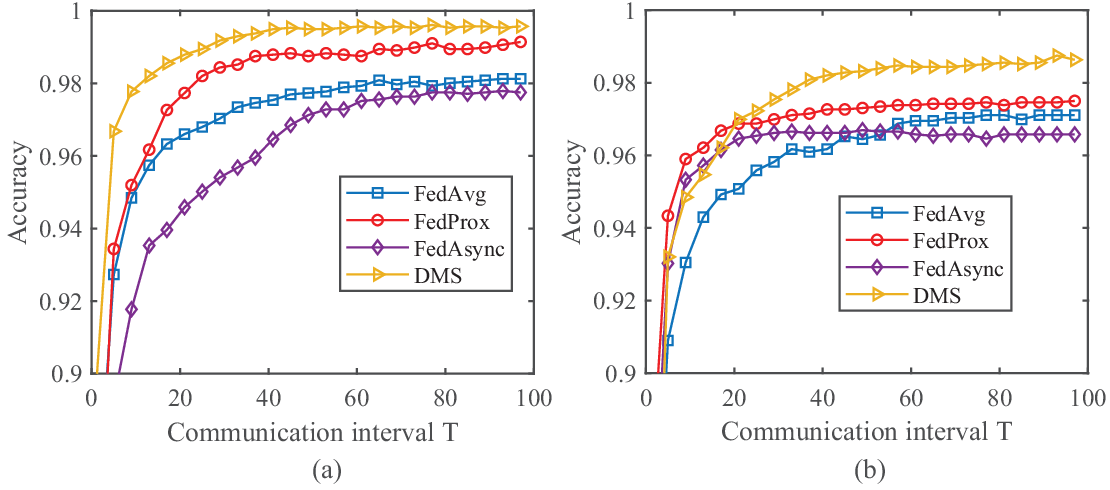}\\
  \caption{Test accuracy versus $T$ for MNIST dataset on (a) $2$-iteration client-selection system and (b) $4$-iteration client-selection system.  }\label{fig_framework} 
\end{figure}

\subsection{Performance of DMS under Dynamic Cases}
Fig.~\ref{fig_dynamic} plots the experimental results of the test accuracy on MNIST, Cifar-10, F-MNIST, and SVHN datasets for various algorithms under Case 3, while Table \ref{tab_dynamic} lists the corresponding highest accuracy of the global models.
First, it is observed that our proposed DMS algorithm outperforms other state-of-the-art algorithms under the dynamic case, demonstrating its effectiveness in improving learning accuracy in practical T-SFL systems with randomly available computing power and data size for each client.
Second, the experimental results show that the DMS algorithm exhibits strong adaptability in practical T-SFL systems.
Specifically, the DMS algorithm achieves a more significant improvement in accuracy under dynamic cases compared to static cases, particularly on the MNIST dataset with the CNN network.
This can be attributed to the dynamic nature of the DMS algorithm, as it continuously adapts to the varying computing power and data size among clients within each communication interval to optimize the weights of the global aggregation.
Third, the experimental results show that the learning accuracy of the \emph{FedAsync} algorithm is lower than that of the static cases.
The \emph{FedAsync} algorithm aggregates the current global model and the global model from adjacent communication intervals, as defined by its generation rule\footnote{The aggregation rule of the \emph{FedAsync} algorithm \cite{DBLP:journals/corr/abs-1903-03934} combines the current global model and the global model from adjacent communication intervals using the formula $\boldsymbol{w}^{t+1} = \gamma \boldsymbol{w}^{t} + (1-\gamma)\frac{1}{N} \sum_{i=1}^N \boldsymbol{w}_i^{t+1}$, where $\gamma$ is the depreciation factor, and in this paper, it is set to $\gamma=0.5$.}.
However, in the T-SFL system with dynamic computing power and data size, the numbers of local iterations on each client vary randomly within each communication interval, leading to significant divergence in the global models of adjacent communication intervals.
Considering the aggregation rule and the practical T-SFL system's characteristics, the \emph{FedAsync} algorithm performs worse in terms of learning accuracy compared to the T-SFL system with fixed computing power and data size.

Fig. \ref{fig_bar} plots the probability of client participation in the DMS algorithm on Cifar-10 under Case 3. The results indicate that the proposed DMS algorithm has a preference for local models with higher computing power, suggesting that the algorithm is more likely to aggregate higher-quality local models. This observation is consistent with the analytical results presented in \textbf{Corollary \ref{coro_1}}.

Fig. \ref{fig_framework} plots the experimental results of the test accuracy on MNIST for different client-selection systems \cite{9237168} under various algorithms.
Table \ref{tab_dynamic} lists the corresponding highest test accuracy of the global models.
In the client-selection system, the local model can only be uploaded by a client if the number of iterations completed by it exceeds the predefined threshold. For example, in Fig. \ref{fig_framework} (a), only those clients who have completed more than two local training iterations are eligible to upload their local models.
It can be observed from Fig. \ref{fig_framework} and Table \ref{tab_dynamic} that the DMS algorithm achieves higher learning accuracy than other algorithms in the client-selection system. This result highlights the effectiveness of the proposed DMS algorithm in improving learning performance in client-selection systems with heterogeneous computing power and data size, compared to conventional aggregation algorithms.
\section{Conclusion}\label{sec_con}

In this paper, we have proposed a T-SFL scheme that enables global model aggregation from diverse clients with varying numbers of local training iterations in a time-driven manner.
Furthermore, we have developed an upper bound on the loss function to evaluate the learning performance of T-SFL.
Using this bound, we have obtained the optimal aggregation weights that minimize the loss function.
Moreover, we have proposed an aggregation algorithm aimed at enhancing the fairness, robustness, security, and learning accuracy of T-SFL.
The experimental results have indicated that our proposed T-SFL framework, when combined with the DMS algorithm, can achieve significant reductions in latency (up to 50\%) and better learning performance compared to benchmark algorithms.  In particular, the DMS algorithm has improved accuracy by up to 6\% on MNIST and 7\% on SVHN when compared to the \emph{FedProx} and \emph{FedAsync} algorithms.

This work suggests potential avenues for future research.
First, conducting a large number of local iterations can lead to overfitting of the local model and waste computing power.
Therefore, optimizing the number of iterations per client within each communication interval remains a challenge.
Second, exploring how to optimize the length of the communication interval is another area that requires further investigation.
Third, the exploration of jointly optimizing aggregation weights while considering both system and data heterogeneity is a promising direction for future study.

\bibliographystyle{IEEEtran}
\bibliography{reference}

\begin{IEEEbiography}[{\includegraphics[width=1in,height=1.25in,clip,keepaspectratio]{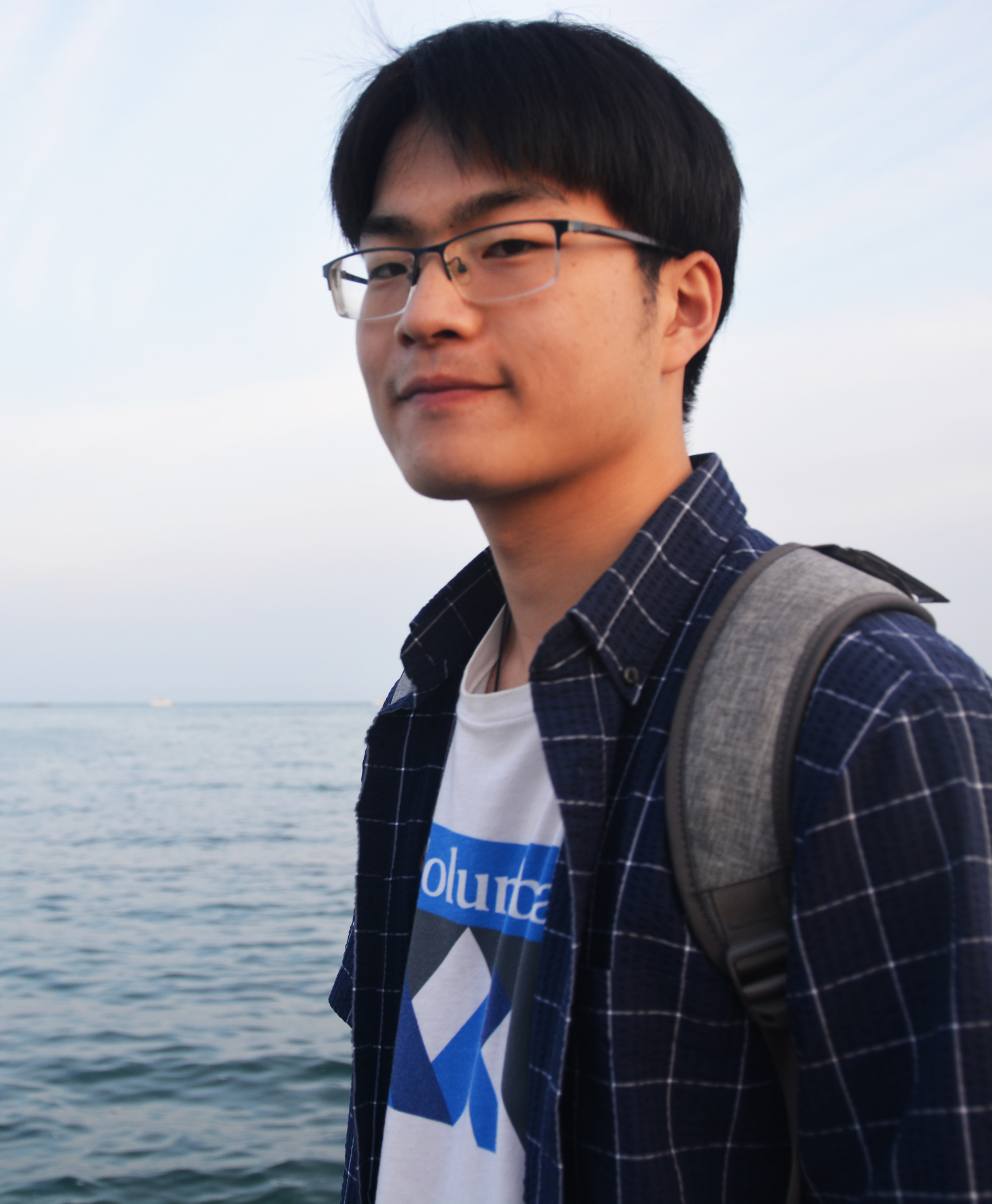}}]
{Yumeng Shao} received the B.S. degree from the School of Electronic and
Optical Engineering, Nanjing University of Science and Technology, Nanjing, China, in 2019, where he is pursuing the Ph.D. degree currently. His research interests include distributed machine learning, blockchain, game theory, and trusted AI.
\end{IEEEbiography}

\begin{IEEEbiography}[{\includegraphics[width=1in,height=1.25in,clip,keepaspectratio]{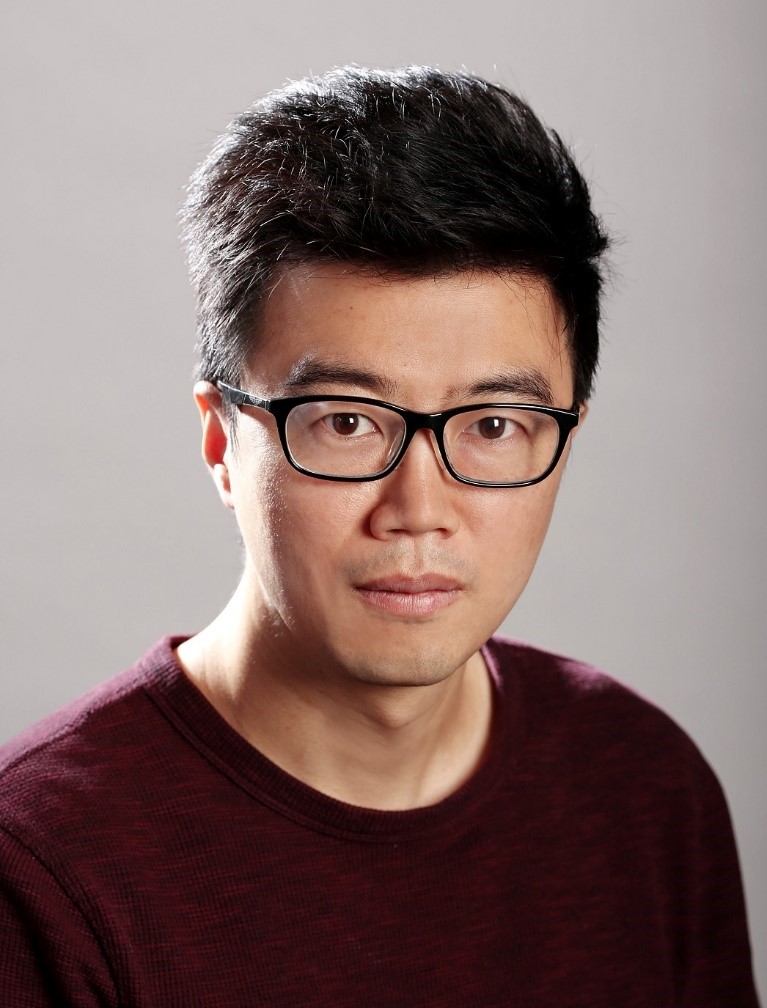}}]
{Jun Li} (M¡¯09-SM¡¯16) received Ph. D degree in Electronic Engineering from Shanghai Jiao Tong University, Shanghai, P. R. China in 2009. From January 2009 to June 2009, he worked in the Department of Research and Innovation, Alcatel Lucent Shanghai Bell as a Research Scientist. From June 2009 to April 2012, he was a Postdoctoral Fellow at the School of Electrical Engineering and Telecommunications, the University of New South Wales, Australia. From April 2012 to June 2015, he was a Research Fellow at the School of Electrical Engineering, the University of Sydney, Australia. From June 2015 to now, he is a Professor at the School of Electronic and Optical Engineering, Nanjing University of Science and Technology, Nanjing, China. He was a visiting professor at Princeton University from 2018 to 2019. His research interests include network information theory, game theory, distributed intelligence, multiple agent reinforcement learning, and their applications in ultra-dense wireless networks, mobile edge computing, network privacy and security, and industrial Internet of things. He has co-authored more than 200 papers in IEEE journals and conferences, and holds 1 US patents and more than 10 Chinese patents in these areas. He was serving as an editor of IEEE Communication Letters and TPC member for several flagship IEEE conferences. He received Exemplary Reviewer of IEEE Transactions on Communications in 2018, and best paper award from IEEE International Conference on 5G for Future Wireless Networks in 2017.
\end{IEEEbiography}

\begin{IEEEbiography}[{\includegraphics[width=1in,height=1.25in,clip,keepaspectratio]{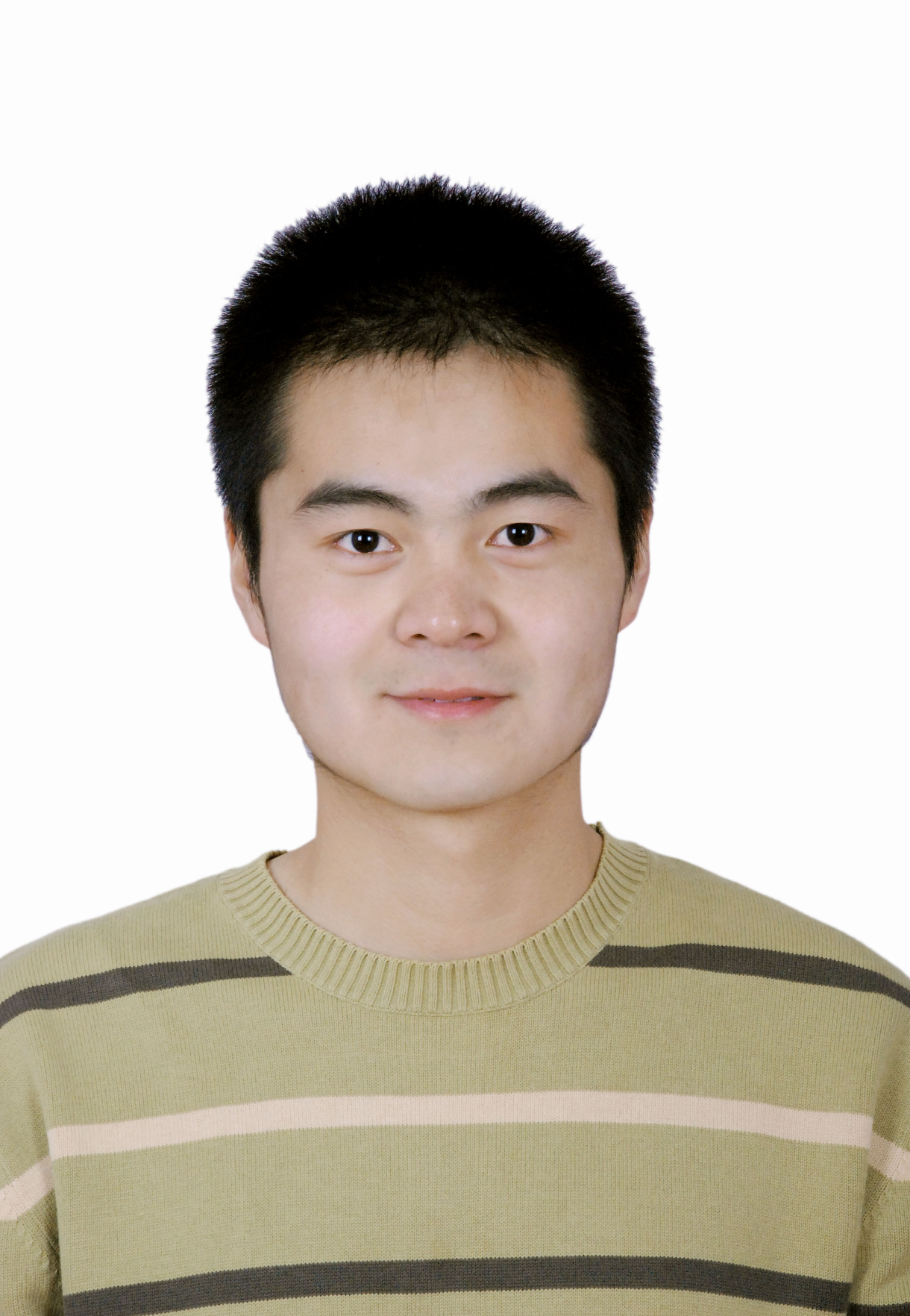}}]
{Long Shi} (S'10-M'15) received the Ph.D. degree in Electrical Engineering from  University of New South Wales, Sydney, Australia, in 2012. From 2013 to 2016, he was a Postdoctoral Fellow at the Institute of Network Coding, Chinese University of Hong Kong, China. From 2014 to 2017, he was a Lecturer at Nanjing University of Aeronautics and Astronautics, Nanjing, China. From 2017 to 2020, he was a Research Fellow at  Singapore University of Technology and Design. Currently, he is a Professor at  School of Electronic and Optical Engineering, Nanjing University of Science and Technology, Nanjing, China. His research interests include blockchain networks, mobile edge computing, and wireless network coding.
\end{IEEEbiography}

\begin{IEEEbiography}[{\includegraphics[width=1in,height=1.25in,clip,keepaspectratio]{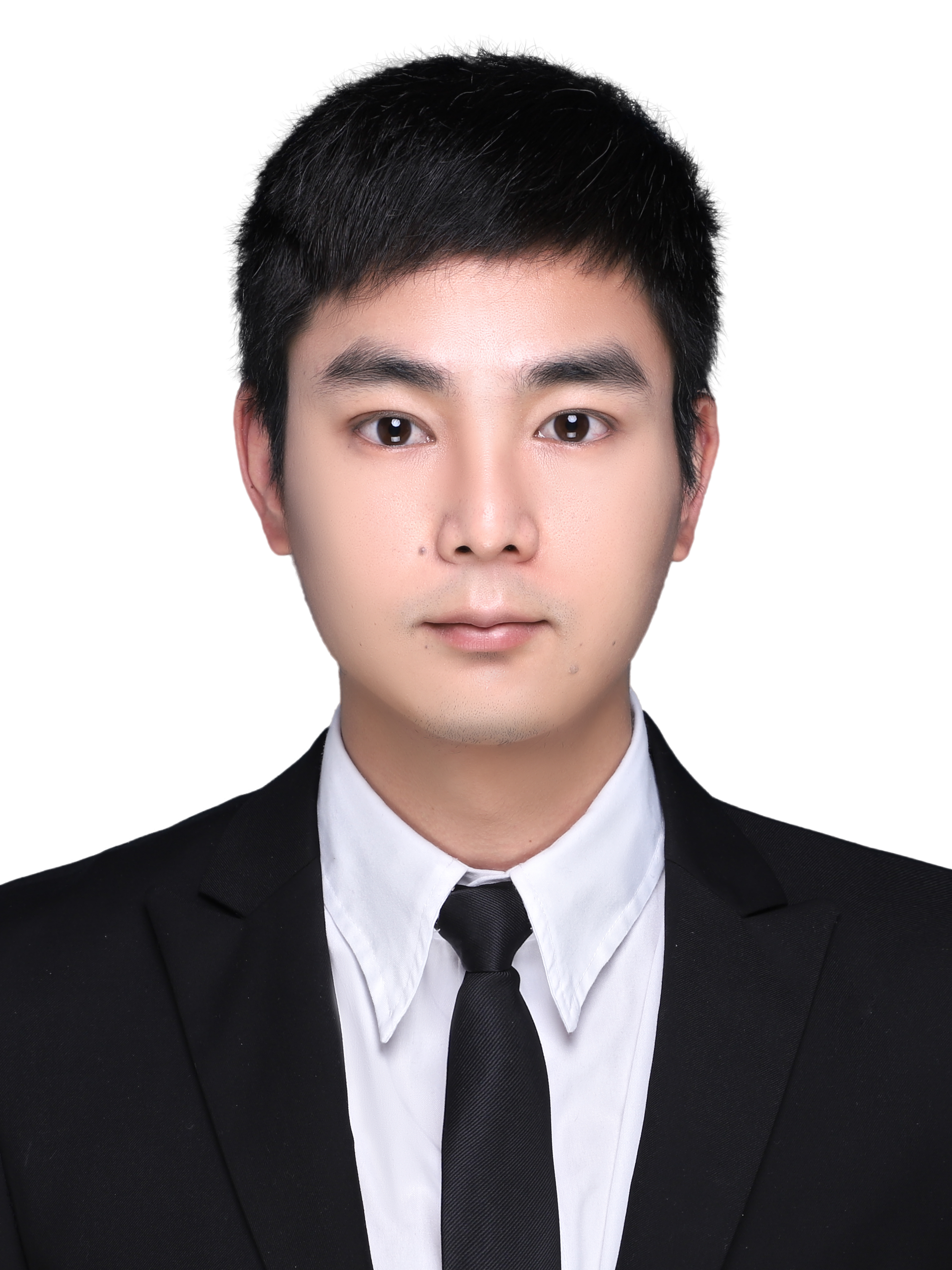}}]
{Kang Wei} received the B.Sc. degree in information engineering from Xidian University, Xi¡¯an, China, in 2014, and the M.Sc. degree from the School of Electronic and
Optical Engineering, Nanjing University of Science and Technology, Nanjing, China, in 2018, where he is currently pursuing the Ph.D. degree. His current research interests include data privacy and security, differential privacy, AI and machine learning, information theory, and channel coding theory in NAND flash memory.
\end{IEEEbiography}

\begin{IEEEbiography}[{\includegraphics[width=1in,height=1.25in,clip,keepaspectratio]{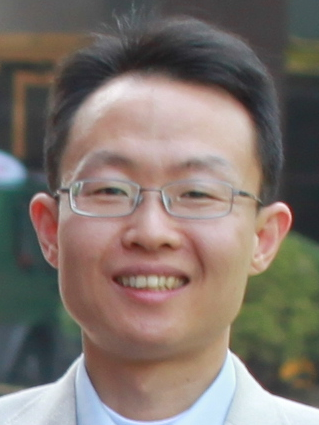}}]
{Ming Ding} (M'12-SM'17) received the B.S. and M.S. degrees (with first-class Hons.) in electronics engineering from Shanghai Jiao Tong University (SJTU), Shanghai, China, and the Doctor of Philosophy (Ph.D.) degree in signal and information processing from SJTU, in 2004, 2007, and 2011, respectively. From April 2007 to September 2014, he worked at Sharp Laboratories of China in Shanghai, China as a Researcher/Senior Researcher/Principal Researcher. Currently, he is a senior research scientist at Data61, CSIRO, in Sydney, NSW, Australia. His research interests include information technology, data privacy and security, and machine learning and AI. He has authored more than 150 papers in IEEE journals and conferences, all in recognized venues, and around 20 3GPP standardization contributions, as well as a book ¡°Multi-point Cooperative Communication Systems: Theory and Applications¡± (Springer, 2013). Also, he holds 21 US patents and has co-invented another 100+ patents on 4G/5G technologies. Currently, he is an editor of \emph{IEEE Transactions on Wireless Communications and IEEE Communications Surveys and Tutorials}. Besides, he has served as a guest editor/co-chair/co-tutor/TPC member for multiple IEEE top-tier journals/conferences and received several awards for his research work and professional services.
\end{IEEEbiography}

\begin{IEEEbiography}[{\includegraphics[width=1in,height=1.25in,clip,keepaspectratio]{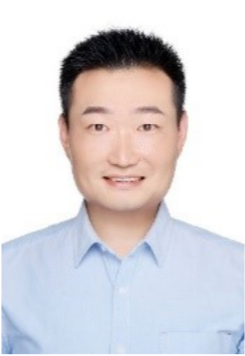}}]
{Qianmu Li}, professor and doctoral supervisor;
foreign academician of the Russian Academy of
Natural Sciences; member of Academic Committee of Nanjing University of technology, director of information construction and Management
Department of Nanjing University of technology;
Vice chairman of Jiangsu Science and Technology Association. Academic member of Cyber
Engineering Laboratory of State Grid, deputy
director of intelligent education special committee of National Computer Basic Teaching and
Research Association, chief expert of e-government platform in Jiangsu
Province, vice chairman of Jiangsu Digital Government Standardization Technical Committee, President of Nanjing Computer Society, vice
president of Jiangsu Cyber Engineering Society, executive director of
Jiangsu Computer Society, Secretary General of Jiangsu Internet Finance Association. He was selected as the first network and security outstanding talent of China communication society, the young and
middle-aged leading talent of Jiangsu Province. He has won more
than ten first and second prizes, including the science and technology
progress award of the Ministry of education, the science and technology
award of Jiangsu Province, the teaching achievement award of Jiangsu
Province, the outstanding achievement award of scientific research in
universities of the Ministry of education, and the best paper awards such
as ISKE, AAAI and ICCC.
\end{IEEEbiography}

\begin{IEEEbiography}[{\includegraphics[width=1in,height=1.25in,clip,keepaspectratio]{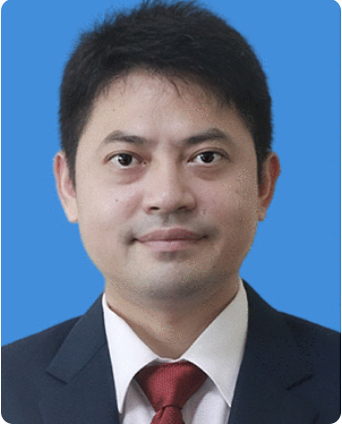}}]
{Zengxiang Li} is Executive Vice President of Digital Research Institute of ENN Group. Dr. Li received his Ph.D Degree in School of Com- puter Science and Engineering Nanyang Tech- nological University in 2012. His Research In- terest Includes, Industrial IoT, Blockchain, AI, Federated Learning, Incentive Mechanism and Privacy-Preserving technology.
\end{IEEEbiography}

\begin{IEEEbiography}[{\includegraphics[width=1in,height=1.25in,clip,keepaspectratio]{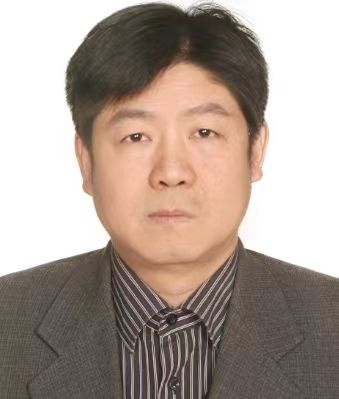}}]
	{Wen Chen} (M¡¯03¨CSM¡¯11) is a tenured Professor with the Department of Electronic Engineering, Shanghai Jiao Tong University, China, where he is the director of Broadband Access Network Laboratory. He is a fellow of Chinese Institute of Electronics and the distinguished lecturers of IEEE Communications Society and IEEE Vehicular Technology Society. He is the Shanghai Chapter Chair of IEEE Vehicular Technology Society, Editors of IEEE Transactions on Wireless Communications, IEEE Transactions on Communications, IEEE Access and IEEE Open Journal of Vehicular Technology. His research interests include multiple access, wireless AI and meta-surface communications. He has published more than 120 papers in IEEE journals and more than 120 papers in IEEE Conferences, with citations more than 8000 in google scholar.
\end{IEEEbiography}
\begin{IEEEbiography}[{\includegraphics[width=1in,height=1.25in,clip,keepaspectratio]{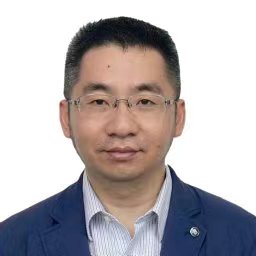}}]
{Shi Jin} received the B.S. degree in communications engineering from the Guilin University of Electronic Technology, Guilin, China, in 1996, the M.S. degree from the Nanjing University of Posts and Telecommunications, Nanjing, China, in 2003, and the Ph.D. degree in information and communications engineering from Southeast University, Nanjing, in 2007. From June 2007 to October 2009, he was a Research Fellow with the Adastral Park Research Campus, University College London, London, U.K. He is currently with the Faculty of the National Mobile Communications Research Laboratory, Southeast University. His research interests include space time wireless communications, random matrix theory, and information theory. He and his coauthors have been awarded the 2011 IEEE Communications Society Stephen O. Rice Prize Paper Award in the field of communication theory and the 2010 Young Author Best Paper Award by the IEEE Signal Processing Society. He serves as an Associate Editor for the IEEE Transactions on Communications, IEEE  Transactions on Wireless Communications, the IEEE Communications Letters, and IET Communications.
\end{IEEEbiography}

\appendices
\clearpage
\section{Proof of \textbf{Theorem \ref{threm_final}}} \label{apd_B}

Since $\Vert \bar{\boldsymbol{w}}^{tH+h} - \boldsymbol{w}^{t}_{i,h} \Vert^2 = \Vert \bar{\boldsymbol{w}}^{tH+h} - \bar{\boldsymbol{w}}^{tH} \Vert^2  +\Vert \boldsymbol{w}^{t}_{i,h} - \bar{\boldsymbol{w}}^{tH} \Vert^2 - 2 \langle\bar{\boldsymbol{w}}^{tH+h} - \bar{\boldsymbol{w}}^{tH}, \boldsymbol{w}^{t}_{i,h} - \bar{\boldsymbol{w}}^{tH}\rangle$ and $- 2 \sum_{i=1}^N \rho^t_i \alpha_{i,h}^t \langle\bar{\boldsymbol{w}}^{tH+h} - \bar{\boldsymbol{w}}^{tH}, \boldsymbol{w}^{t}_{i,h} - \bar{\boldsymbol{w}}^{tH} \rangle =  -2 \Vert \bar{\boldsymbol{w}}^{tH+h} -\bar{\boldsymbol{w}}^{tH} \Vert^2$, we obtain

\begin{align}
\sum_{i=1}^N \rho^t_i \Vert \bar{\boldsymbol{w}}^{tH+h} - \boldsymbol{w}^{t}_{i,h} \Vert^2
\leq  \sum_{i=1}^N \rho^t_i \Vert \boldsymbol{w}^{t}_{i,h} - \bar{\boldsymbol{w}}^{tH} \Vert^2.
\end{align}
Then, we have
\begin{align}
&\sum_{i=1}^N \rho^t_i \alpha_{i,h}^t \Vert \bar{\boldsymbol{w}}^{tH+h} - \boldsymbol{w}^{t}_{i,h} \Vert^2
=  \sum_{i=1}^N \rho^t_i \alpha_{i,h}^t \Vert \eta \sum_{j=0}^{h-1} \boldsymbol{g}^t_{i,j} \alpha^t_{i,j} \Vert^2 \leq \nonumber \\
& \mathbb{E}_\xi[\eta^2 h \sum_{i=1}^N \rho^t_i \alpha_{i,h}^t G^2 \sum_{j=1}^{h-1}  \alpha^t_{i,j} \leq (H-1)\eta^2 G^2 \sum_{i=1}^N \rho^t_i \alpha_{i,h}^t \tau_i^t.
\end{align}

Since that $\bar{\boldsymbol{g}}^{t}_{i,h} = \mathbb{E}_\xi [\boldsymbol{g}^{t}_{i,h}]$ and $\mathbb{E}_\xi[\Vert(\boldsymbol{g}^{t}_{i,h}-\bar{\boldsymbol{g}}^{t}_{i,h})(\boldsymbol{g}^{t}_{j,h}-\bar{\boldsymbol{g}}^{t}_{j,h}) \Vert |_{i \neq j}] = 0$, we have

\begin{align}\label{eq_A1+A2}
&\mathbb{E}_\xi[\Vert \bar{\boldsymbol{w}}^{tH+h+1} - \boldsymbol{w}^* \Vert^2]= \notag \\
& \mathbb{E}_\xi[\underbrace{\Vert \bar{\boldsymbol{w}}^{tH+h} -\eta \sum_{i=1}^N \rho^t_i \bar{\boldsymbol{g}}_{i,h}^t \alpha_{i,h}^t - \boldsymbol{w}^* \Vert^2}_{U_1}] \notag \\
&+ \eta^2 \mathbb{E}_\xi[\Vert  \sum_{i=1}^N \rho^t_i \alpha_{i,h}^t (\bar{\boldsymbol{g}}_{i,h}^t - \boldsymbol{g}_{i,h}^t) \Vert^2] + \mathbb{E}_\xi[U_2],
\end{align}
where $U_2 = 2 \eta \langle \bar{\boldsymbol{w}}^{tH+h} -\eta \sum_{i=1}^N \rho^t_i \bar{\boldsymbol{g}}_{i,h}^t \alpha_{i,h}^t - \boldsymbol{w}^*,$ $\sum_{i=1}^N \rho^t_i \alpha_{i,h}^t (\bar{\boldsymbol{g}}_{i,h}^t - \boldsymbol{g}_{i,h}^t)\rangle = 0$.
Plugging $\mathbb{E}_\xi[U_2]=0$ into (\ref{eq_A1+A2}), we have

\begin{align}
\mathbb{E}_\xi[\Vert \bar{\boldsymbol{w}}^{tH+h+1} - \boldsymbol{w}^* \Vert^2]
\leq  \mathbb{E}_\xi[U_1] + \eta^2 N \sum_{i=1}^N \alpha_{i,h}^t (\rho^t_i)^2 \sigma_i^2.
\end{align}

Since $F_i(\cdot)$ is $L$-smooth in expectation and $\nabla F_i(\boldsymbol{w}_i^*)=0$ (i.e., $\Vert \bar{\boldsymbol{g}}_{i,h}^t \Vert^2 \leq 2L [F_i(\boldsymbol{w}_{i,h}^t) - F_i(\boldsymbol{w}_i^*)]$), we have

\begin{align}
U_1 = & \eta^2 \Vert \sum_{i=1}^N \rho^t_i \bar{\boldsymbol{g}}_{i,h}^t \alpha_{i,h}^t \Vert^2 +\Vert \bar{\boldsymbol{w}}^{tH+h} - \boldsymbol{w}^* \Vert^2 \notag \\
& -2\eta \langle\bar{\boldsymbol{w}}^{tH+h} - \boldsymbol{w}^*, \sum_{i=1}^N \rho^t_i \bar{\boldsymbol{g}}_{i,h}^t \alpha_{i,h}^t\rangle, \notag \\
\leq & 2 \eta^2 L N \sum_{i=1}^N (\rho^t_i)^2 \alpha_{i,h}^t [F_i(\boldsymbol{w}_{i,h}^t) - F_i(\boldsymbol{w}_i^*)] \notag \\
& + \Vert \bar{\boldsymbol{w}}^{tH+h} - \boldsymbol{w}^* \Vert^2  + V_1 +V_2,
\end{align}
where $V_1 = -2\eta \sum_{i=1}^N \rho^t_i \langle\bar{\boldsymbol{w}}^{tH+h} - \boldsymbol{w}^t_{i,h},  \bar{\boldsymbol{g}}_{i,h}^t \alpha_{i,h}^t\rangle$ and $V_2=-2\eta \sum_{i=1}^N \rho^t_i \langle \boldsymbol{w}^t_{i,h} - \boldsymbol{w}^*,  \bar{\boldsymbol{g}}_{i,h}^t \alpha_{i,h}^t\rangle$.

Since $F_i(\cdot)$ is $L$-smooth and $F_i(\bar{\boldsymbol{w}}^{tH+h} ) - F_i(\boldsymbol{w}_{i,h}^t) \geq F_i(\boldsymbol{w}_{i,h}^t) - F_i(\boldsymbol{w}_i^*)$, we have

\begin{align}
V_1 \leq & - 2 \eta \sum_{i=1}^N \rho_i^t \alpha_{i,h}^t [ F_i(\boldsymbol{w}_{i,h}^t) - F_i(\boldsymbol{w}_i^*) ] \notag \\
& + \eta L \sum_{i=1}^N \rho_i^t \alpha_{i,h}^t \Vert \boldsymbol{w}_{i,h}^t - \bar{\boldsymbol{w}}^{tH+h} \Vert^2.
\end{align}

According to the optimization theory, we have $\langle \boldsymbol{w}^t_{i,h} - \boldsymbol{w}_i^*,  \bar{\boldsymbol{g}}_{i,h}^t\rangle \geq 0$ and

\begin{align}
V_2 \leq  \sum_{i=1}^N \rho^t_i \alpha_{i,h}^t \Vert \boldsymbol{w}_i^* - \boldsymbol{w}^* \Vert^2 + \eta^2 \sum_{i=1}^N \rho^t_i \alpha_{i,h}^t \Vert \bar{\boldsymbol{g}}_{i,h}^t \Vert^2.
\end{align}

Recall that $\rho^t_i N \leq \theta$, we have

\begin{align}
& \mathbb{E}_\xi[\Vert \bar{\boldsymbol{w}}^{tH+h+1} - \boldsymbol{w}^* \Vert^2] \leq \notag \\
& \mathbb{E}_\xi[\Vert \bar{\boldsymbol{w}}^{tH+h} - \boldsymbol{w}^* \Vert^2] + \eta^2 \sum_{i=1}^N \rho^t_i \alpha^t_{i,h}  \Vert \boldsymbol{w}_i^* - \boldsymbol{w}^* \Vert^2 \nonumber \\
&+ 2 \eta [\eta L (1+\theta) - 1] \sum_{i=1}^N \rho^t_i \alpha_{i,h}^t [F_i(\boldsymbol{w}_{i,h}^t) - F_i(\boldsymbol{w}_i^*)] \notag \\
&+ \eta^2 N \sum_{i=1}^N (\rho^t_i)^2 \alpha_{i,h}^t \sigma_i^2 + \eta^3 L (H-1) G^2 \sum_{i=1}^N \rho^t_i \alpha_{i,h}^t \tau_i^t.
\end{align}

Based on the rule that the sum of expectations is equal to the expectation of the sum and $\Vert \boldsymbol{w}_{l}^t - \boldsymbol{w}_i^* \Vert^2 \leq  \Vert \boldsymbol{w}^T - \boldsymbol{w}^* + \boldsymbol{w}^* - \boldsymbol{w}_i^* \Vert^2 \leq 2 \Vert \boldsymbol{w}^T - \boldsymbol{w}^* \Vert^2 + 2 \Vert \boldsymbol{w}^* - \boldsymbol{w}_i^* \Vert^2$, we sum the left and right items with respect to both $h$ and $t$, and obtain

\begin{align}
&  \mathbb{E}_i [\Vert \bar{\boldsymbol{w}}^{TH} - \boldsymbol{w}^* \Vert^2 ] \leq \notag \\
& \mathbb{E}_i [\Vert \bar{\boldsymbol{w}}^0 - \boldsymbol{w}^* \Vert^2 ]
+ \eta^2 N \sum_{t=0}^{T-1} \sum_{i=1}^N (\rho^t_i)^2 \tau_i^t \sigma_i^2 \notag \\
& + \eta^3 L (H-1) G^2 \sum_{i=1}^N \sum_{t=0}^{T-1} \rho^t_i(\tau_i^t)^2 \notag \\
& + 2 \eta L [\eta L (1+\theta) - 1] \sum_{i=1}^N \sum_{t=0}^{T-1} \rho^t_i \tau_i^t \Vert \boldsymbol{w}^T - \boldsymbol{w}^* \Vert^2\nonumber \\
& +(2 \eta L (\eta L (1+\theta) - 1) +\eta^2) \sum_{i=1}^N \sum_{t=0}^{T-1} \rho^t_i \tau_i^t \Gamma]_i.
\end{align}

Thus, we have

\begin{align}
& \Vert \boldsymbol{w}^{T} - \boldsymbol{w}^* \Vert^2 \leq
\frac{\Vert \bar{\boldsymbol{w}}^0 - \boldsymbol{w}^* \Vert^2 + \eta^2 N  \sum_{i=1}^N \sigma_i^2 \sum_{t=0}^{T-1} (\rho^t_i)^2 \tau_i^t  }{1+ 2 \eta L [1- \eta L (1+\theta)] \sum_{i=1}^N \sum_{t=0}^{T-1} \rho^t_i \tau_i^t} \nonumber \\
& +\frac{ \eta^3 L (H-1) G^2 \sum_{i=1}^N \sum_{t=0}^{T-1} \rho^t_i(\tau_i^t)^2 }{1+ 2 \eta L [1 - \eta L (1+\theta)] \sum_{i=1}^N \sum_{t=0}^{T-1} \rho^t_i \tau_i^t} \notag \\
&+ \frac{ (2 \eta L (\eta L (1+\theta) - 1) +\eta^2) \sum_{i=1}^N \sum_{t=0}^{T-1} \rho^t_i \tau_i^t \Gamma_i}{1 + 2 \eta L [1 - \eta L (1+\theta) ] \sum_{i=1}^N \sum_{t=0}^{T-1} \rho^t_i \tau_i^t}.
\end{align}

Finally, since $F(\cdot)$ is a linear combination of $F_i(\cdot)$, $L$-smooth and still holds in expectation. We have
\begin{equation}
\Vert \nabla F(\boldsymbol{w}^T) \Vert^2 \leq 2 L [F(\boldsymbol{w}^T) - F(\boldsymbol{w}^*)] \leq L^2 \Vert \boldsymbol{w}^T - \boldsymbol{w}^* \Vert^2,
\end{equation}
which completes the proof.

\section{Proof of \textbf{Theorem \ref{theo_rho}}}\label{apd_rho}
Define a function $g(\rho_i^t)$ as the upper bound of the loss function in (\ref{eq_4}), i.e.,
\begin{align}
g(\rho_i^t) = L^2  \frac{\Vert \boldsymbol{w}^0 - \boldsymbol{w}^* \Vert^2 +X +Y + Z }{W}.
\end{align}

Given that the derivative of the extreme point is $0$, i.e., $\frac{\mathrm{d}(g(\rho_i^*))}{\mathrm{d}\rho_i^*}= 0$, we have
\begin{align}
&\frac{\mathrm{d}(g(\rho_i^t))}{\mathrm{d}\rho_i^t} = L^2 (1 + B \sum_{i=1}^N \sum_{t=0}^{T-1} \rho_i^* \tau_i^t) (2 \eta^2 N \sigma_i^2) \rho_i^* \tau_i^t \notag \\
&- L^2 \left[ A \Gamma_i \tau_i^t + C(\tau_i^t)^2 \right]
\cdot (1 + B \sum_{i=1}^N \sum_{t=0}^{T-1} \rho_i^* \tau_i^t) \notag \\
& - L^2 B \tau_i^t \left[ \Vert \boldsymbol{w}^0 - \boldsymbol{w}^* \Vert^2 +\eta^2 N \sum_{i=1}^N \sum_{t=0}^{T-1} \sigma_i^2 (\rho_i^*)^2 \tau_i^t \right. \notag \\
& \left. + A \sum_{i=1}^N \sum_{t=0}^{T-1} \Gamma_i \rho_i^* \tau_i^t + C \sum_{i=1}^N \sum_{t=0}^{T-1} \rho_i^* (\tau_i^t)^2 \right] = 0,
\end{align}

where
\begin{align}
&A= \eta \{ 2 L[\eta L (1+\theta) - 1 ] + \eta \} , \nonumber  \\
&B = 2 \eta L [1 - \eta L (1+ \theta)], \nonumber \\
&C = \eta^3 L (H-1) G^2.
\end{align}

Note that the coefficient of the primary term is greater than $0$, i.e., $\frac{L}{2} (1 + B \sum_{i=1}^N \sum_{t=0}^{T-1} \rho_i^* \tau_i^t) (2 \eta N \sigma_i^2) \tau_i^t > 0$, we conclude that the extreme point is the minimum point of $g(\rho_i^t)$. Therefore, we obtain the optimal point of $\rho_i^t$ as
\begin{align}\label{eq_opt_rho}
  \rho_i^* = \frac{B D + ( A \Gamma_i  + C \tau_i^t )  (1 + B \sum_{i=1}^N \sum_{t=0}^{T-1} \rho_i^* \tau_i^t) }{(1 + B \sum_{i=1}^N \sum_{t=0}^{T-1} \rho_i^* \tau_i^t)  (2 \eta^2 N \sigma_i^2) },
\end{align}
where
\begin{align}
D = & \Vert \boldsymbol{w}^0 - \boldsymbol{w}^* \Vert^2 +\eta^2 N \sigma_i^2 \sum_{i=1}^N \sum_{t=0}^{T-1} (\rho_i^*)^2 \tau_i^t \notag \\
&+ A \Gamma_i \sum_{i=1}^N \sum_{t=0}^{T-1} \rho_i^* \tau_i^t + C \sum_{i=1}^N \sum_{t=0}^{T-1} \rho_i^* (\tau_i^t)^2.
\end{align}

According to the Nash equilibrium theory \cite{DBLP:journals/candie/HouZY20}, we find that the items $\sum_{i=1}^N \sum_{t=0}^{T-1} (\rho_i^*)^2 \tau_i^t$, $\sum_{i=1}^N \sum_{t=0}^{T-1} \rho_i^* \tau_i^t$ and $\sum_{i=1}^N \sum_{t=0}^{T-1} \rho_i^* (\tau_i^t)^2$ are constants for an optimal system. Therefore, we confirm that (\ref{eq_opt_rho}) is a closed-form of $\rho_i^*$, which completes the proof.

\section{Proof of \textbf{Theorem \ref{theorem_5}}} \label{app_E}

Since $F(\cdot)$ is $L$-smooth and from the definition of $\bar{\boldsymbol{w}}$, we have
\begin{align}
&F(\bar{\boldsymbol{w}}^{tH+h+1}) - F(\bar{\boldsymbol{w}}^{tH+h})\leq \notag \\
& \langle \nabla F (\bar{\boldsymbol{w}}^{tH+h}),- \eta \sum_{i=1}^N \rho_i^t \alpha_{i,h}^t \bar{\boldsymbol{g}}_{i,h}^t \rangle \notag \\
&+ \frac{L}{2} \Vert \bar{\boldsymbol{w}}^{tH+h+1} - \bar{\boldsymbol{w}}^{tH+h} \Vert^2 \leq \notag \\
& - \eta \nabla F (\bar{\boldsymbol{w}}^{tH+h})^\top \mathbb{E}_i [ \nabla F_i (\bar{\boldsymbol{w}}^{tH+h}) ] \notag\\
&+ \frac{\eta^2 L}{2} \mathbb{E}_i [ \Vert \nabla F_i (\bar{\boldsymbol{w}}^{tH+h}) \Vert^2].
\end{align}
According to \textbf{Definition \ref{defin_2}} and \textbf{Assumption \ref{assumption_nonconvex}}, we have

\begin{align}
&F(\bar{\boldsymbol{w}}^{tH+h+1}) - F(\bar{\boldsymbol{w}}^{tH+h}) \notag \\
\leq & - \eta (\epsilon - \frac{\eta L   V^2}{2}) \Vert \nabla F (\bar{\boldsymbol{w}}^{tH+h}) \Vert^2.
\end{align}
Then, we derive that
\begin{align}
F(\boldsymbol{w}^{T}) - & F(\boldsymbol{w}^{0}) = \sum_{t=0}^{T-1}\sum_{h=0}^{H-1} \left[ F(\bar{\boldsymbol{w}}^{tH+h+1}) - F(\bar{\boldsymbol{w}}^{tH+h}) \right] \notag \\
\leq & - \eta (\epsilon - \frac{\eta L   V^2}{2}) \sum_{t=0}^{T-1}\sum_{h=0}^{H-1} \Vert \nabla F (\bar{\boldsymbol{w}}^{tH+h}) \Vert^2.
\end{align}

Since $F_{\mathrm{min}}=F(\boldsymbol{w}^{*})\leq F(\boldsymbol{w}^{T})$, we have
\begin{equation}
  \eta (\epsilon - \frac{\eta L   V^2}{2}) \sum_{t=0}^{T-1} \sum_{h=0}^{H-1} \Vert \nabla F (\bar{\boldsymbol{w}}^{tH+h}) \Vert^2 \leq  F(\boldsymbol{w}^{0}) - F(\boldsymbol{w}^{*}).
\end{equation}
Since $\eta<\frac{2 \epsilon}{V^2   L}$ and $\epsilon - \frac{\eta L   V^2}{2}>0$, we have
\begin{align}
&\eta (\epsilon - \frac{\eta L   V^2}{2}) \sum_{t=0}^{T-1}  \Vert \nabla F (\boldsymbol{w}^{t}) \Vert^2 \notag \\
\leq & \eta (\epsilon - \frac{\eta L   V^2}{2}) \sum_{t=0}^{T-1} \sum_{h=0}^{H-1} \Vert \nabla F (\bar{\boldsymbol{w}}^{tH+h}) \Vert^2 \notag \\
\leq & F(\boldsymbol{w}^{0}) - F(\boldsymbol{w}^{*}).
\end{align}
Finally, we have
  \begin{equation}
    \frac{1}{T} \sum_{t=0}^{T-1} \Vert \nabla F(\boldsymbol{w}^t) \Vert^2 \leq \frac{1}{T} \cdot \frac{2 \eta}{2 \epsilon - \eta L   V^2} \cdot  [F(\boldsymbol{w}^0) - F(\boldsymbol{w}^*)],
  \end{equation}
  which completes the proof.

\vfill

\end{document}